\documentclass[sigconf]{acmart}
\makeatletter
\@ifpackageloaded{xcolor}{}{\usepackage{xcolor}}
\makeatother
\usepackage{stfloats}
\AtBeginDocument{%
  }

\setcopyright{acmlicensed}
\copyrightyear{2026}
\acmYear{2026}
\setcopyright{cc}
\setcctype{by}
\acmConference[CHI '26]{Proceedings of the 2026 CHI Conference on Human Factors in Computing Systems}{April 13--17, 2026}{Barcelona, Spain}
\acmBooktitle{Proceedings of the 2026 CHI Conference on Human Factors in Computing Systems (CHI '26), April 13--17, 2026, Barcelona, Spain}
\acmDOI{10.1145/3772318.3790481}
\acmISBN{979-8-4007-2278-3/2026/04}

\usepackage[nolist,nohyperlinks]{acronym}
\usepackage[acronym]{glossaries}

\usepackage{soul}
\usepackage{framed}

\usepackage{multirow}
\usepackage{makecell} 
\usepackage{subcaption}
\usepackage{pifont} 
\usepackage{csquotes}

\begin{document}
\definecolor{customyellow}{RGB}{255, 220, 74}
\definecolor{customred}{RGB}{255, 198, 198}
\definecolor{customblue}{RGB}{198, 220, 255}
\definecolor{customgreen}{RGB}{44, 160, 44}
\definecolor{custompurple}{RGB}{148, 103, 189}
\definecolor{customblue2}{RGB}{140, 156, 218}

\begin{acronym}
\acro{PwD}{People with Disabilities}
\acro{LLM}{Large Language Models}
\acro{GenAI}{Generative Artificial Intelligence}
\acro{AI}{Artificial Intelligence}
\acro{ADL}{Activities of Daily Living}
\acro{VLM}{Vision Language Models}
\acro{PARs}{Physically Assistive Robots}
\acro{ADA}{Assistive Dexterous Arm}
\acro{HRI}{Human-Robot Interaction}
\acro{PD}{Participatory Design}
\end{acronym}

\newcommand{\cmark}{\ding{51}} 
\newcommand{\xmark}{\ding{55}} 
\newcommand{\umark}{--}
\title{Robot-Assisted Social Dining as a White Glove Service}

\author{Atharva S Kashyap}
\affiliation{%
 \department{Robotics}
 \institution{University of Michigan}
 \city{Ann Arbor}
 \state{Michigan}
 \country{USA}}
\email{katharva@umich.edu}

\author{Ugne Aleksandra Morkute}
\affiliation{%
 \institution{Leiden University}
 \city{Leiden}
 \country{Netherlands}}
\email{ugnea@umich.edu}

\author{Patricia Alves-Oliveira}
\affiliation{%
\department{Robotics}
 \institution{University of Michigan}
 \city{Ann Arbor}
 \state{Michigan}
 \country{USA}}
\email{robopati@umich.edu}

\renewcommand{\shortauthors}{Kashyap et al.}

\begin{abstract}  
  Robot-assisted feeding enables people with disabilities who require assistance eating to enjoy a meal independently and with dignity. However, existing systems have only been tested in-lab or in-home, leaving in-the-wild social dining contexts (e.g., restaurants) largely unexplored. Designing a robot for such contexts presents unique challenges, such as dynamic and unsupervised dining environments that a robot needs to account for and respond to. Through speculative participatory design with people with disabilities, supported by semi-structured interviews and a custom AI-based visual storyboarding tool, we uncovered ideal scenarios for in-the-wild social dining. Our key insight suggests that such systems should: embody the principles of a \textit{white glove service} where the robot (1)~supports multimodal inputs and unobtrusive outputs; (2)~has contextually sensitive social behavior and prioritizes the user; (3)~has expanded roles beyond feeding; (4)~adapts to other relationships at the dining table. Our work has implications for in-the-wild and group contexts of robot-assisted feeding.
\end{abstract}




\begin{teaserfigure}
  \includegraphics[width=\textwidth]{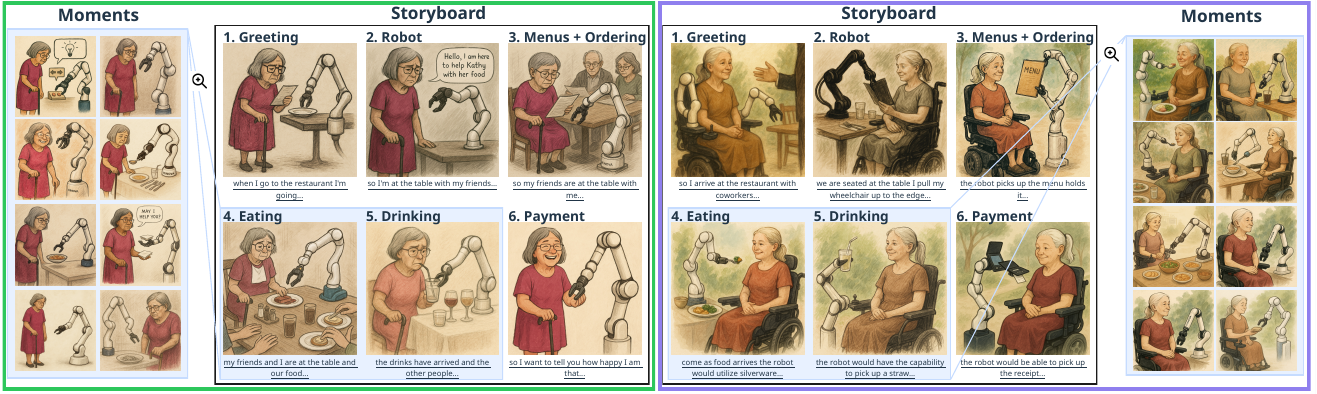}
  \caption{Examples of two Storyboards and their corresponding Moments created using \textsc{Speak2Scene} by P2 (left) and P3 (right).}
  \Description[Participant-created storyboards and moments]{This figure captures P2 and P3's storyboards and moments that are created using Speak2Scene tool.}
  \label{fig:storyboard-ex}
\end{teaserfigure}


\maketitle

\section{Introduction}
We often eat surrounded by other people: bonding with our friends over dinner at a favorite restaurant, connecting with family during breakfast at a cozy cafe, or establishing business relationships over lunch at a corner bistro. These contexts of Social Dining (i.e., to eat with/in company~\cite{design-principles-2023}) are deeply cultural and shape how enjoyable food can feel. Yet, \ac{PwD}\footnote{For simplicity, in this paper, we will use the term ``Person/People with Disabilities''(PwD) to describe the target population of this research, who are people with physical disabilities that affect their ability to use their upper limbs to a point where they require varying levels caregiver support to independently eat.} are often excluded from these experiences. If you recall your own dining experiences at restaurants or cafes, how many times have you seen someone in a wheelchair? The general answer, unfortunately, is: not often. \ac{PwD} frequently report having extremely negative experiences of social dining, sometimes avoiding eating out altogether due to the many barriers they face, such as their perceived burden of depending on a caregiver to eat and frequently having to interrupt conversations at the table to ask for food~\cite{design-principles-2023}. As one participant shared, she often chooses not to eat when she dines with friends at a restaurant, noting: \textit{``a lot of times I take home food''} and \textit{``usually get just to drink''}~(P5). Such experiences illustrate the depth of exclusion that \ac{PwD} feel when dining socially in public. 


Robot-assisted feeding systems~\cite{Feast-2025, lessons-learned-2025} have shown promise in supporting \textit{solitary dining} for \ac{PwD} and several studies addressed the technical performance and safety of these systems. Robot-assisted \textit{social dining}, however, introduces challenges beyond robot-assisted \textit{solitary dining}, which include being able to sustain a conversation while eating, engaging in shared table activities (e.g., toast), and using socially appropriate utensils when dining at a restaurant (e.g., knife and fork for a steak). Some of these challenges are beginning to be addressed. For instance, prior work has studied mealtime assistance for in-the-wild contexts (e.g., homes, cafeterias) with a care recipient~\cite{lessons-learned-2025}, while other researchers have explored personalization through language-based modifications to behavior trees and tested with care recipients at home~\cite{Feast-2025}. Despite these advances, there is a research gap in understanding the needs of \ac{PwD} from the robot-assisted feeding system when they operate in out-of-the-home and out-of-the-lab environments (e.g., restaurants, cafes). Context, dignity, timing, and interpersonal dynamics are central to dining~\cite{systematic-review-2025}, and these factors take on unique significance in social dining contexts, making this an important gap to address.

This paper presents a speculative \ac{PD} study of robot-assisted social dining in public spaces. Our participants are \ac{PwD} with a lived experience of requiring assistance to eat. To conduct the speculative \ac{PD} sessions, we used a custom-made \ac{GenAI}-powered storyboarding tool (see Figure~\ref{fig:storyboard-ex}), followed by semi-structured interviews, to capture the views and perspectives of our participants in their ideal social dining scenarios. The main insight from this work suggests that successful robot-assisted feeding should embody the principles of \textit{white glove service}. In restaurants, \textit{white glove service} is a personalized and detail-oriented approach to provide hospitality to customers~\cite{linkedin-wgs}, which we believe everyone deserves. In line with the CHI 2026 theme of \textit{``Creating Tomorrow Together''}, our study employs speculative and participatory approaches to envision inclusive futures in robot-assisted social dining. Our work sets the stage for future research on robot-assisted feeding for social dining in public spaces, led by insights from \ac{PwD}, who will be the end users of such robot systems. This paper addresses the following research questions:
\begin{itemize}
    \item \textbf{RQ1: \textit{What type of human-robot communication supports \ac{PwD} during social dining?}}
    \item \textbf{RQ2: \textit{How should the robot behavior adapt across different social dining scenarios?}}
    \item \textbf{RQ3: \textit{What should the robot's role be during and outside of mealtime assistance?}}
\end{itemize}

\section{Related Work}
In this section, we present prior work conducted around designing robots by \ac{PwD}, how \ac{GenAI} and speculative design have been used to design robots and assistive technologies, and research advances in assistive robots for dining. 

\subsection{Design of Robots by People with Disabilities}
Designing assistive technologies for \ac{PwD} requires inclusive methodologies that prioritize \ac{PD}, co-design, and functional prototyping. Using \ac{PD} and co-design methodologies to create assistive technologies can help researchers uncover user needs, address unique challenges, and empathize with \ac{PwD}~\cite{pd-2021-reflections}.

Several studies have engaged \ac{PwD} in designing assistive robots. A participatory design study by Azenkot et al.~\cite{azenkot-2016} worked with designers with vision disabilities to come up with design features for service robots in large buildings. Motahar et al.~\cite{motahar-2019} conducted multiple studies with older adults in developing nations using low fidelity prototypes and to create a low-cost robot for medication support. A needs assessment study with older adults by Beer et al.~\cite{beer-2012} provided design recommendations for creating mobile manipulator robots to support aging in-place.

Beyond design of broader assistive robots, several studies have specifically examined robot-assisted feeding. A contextual inquiry study by Bhattacharjee et al.~\cite{bhattacharjee-2019-design} was conducted with an assisted-living community to develop a design framework to build robot-assisted feeding systems that focused on solitary dining. The design framework was then used to quantitatively and qualitatively evaluate three feeding systems to identify areas of improvement~\cite{bhattacharjee-2019-design}. A different study by Nanavati et al.~\cite{design-principles-2023} was conducted to explore the social dimensions of robot-assisted feeding using community-based participatory research with \ac{PwD} using speculative video scenarios and follow-up interviews. In this study, participants were shown conceptual videos depicting robot-assisted feeding in various real-life social dining contexts, such as meals with friends, allowing them to envision and critique possible futures without needing to interact with a working prototype. The insights were distilled into nine design principles intended to inform how the feeding component of robot-assisted systems can be developed and adapted for social dining contexts.~\cite{design-principles-2023}. Furthermore, by applying an iterative user-centered design process, researchers achieved mechanical design enhancements to passive dining devices, such as lower required arm elevation and increased safety through the integration of compliant utensil attachments~\cite{design-improvement-2020}. 

Ljungblad~\cite{Ljungblad}, for instance, argues that assistive feeding is a social and aesthetic experience rather than merely a technical task, and that design-oriented thinking can reveal new possibilities for making feeding robots more acceptable and meaningful. However, a research gap remains in understanding the needs of \ac{PwD} from robots in social dining contexts that extend beyond the act of \textit{feeding}. While prior work proposed design principles for the \textit{feeding} process focused on solitary dining~\cite{bhattacharjee-2019-design} and social dining~\cite{design-principles-2023}, little research has explored the broader needs of \ac{PwD} from robot-assisted social dining in restaurant settings, which we address in this paper.   

\subsection{GenAI to Design Robots}
Recent work has explored \ac{GenAI}, such as \ac{LLM} and \ac{VLM} as tools for designing and customizing \ac{HRI}. These tools allow users to express preferences and initiate behaviors~\cite{initiate-2024}, or modify robot configurations using natural language and multi-modal inputs~\cite{robot-manipulation-2023}, offering novel opportunities for \ac{PD}.

For instance, RoboCrafter-QA~\cite{robo-crafter-2025} uses \ac{LLM} as evaluators to compare soft robot designs based on language-described tasks, offering a promising step toward integrating natural language understanding into the robot design process. However, the tool was not evaluated with human users; it was tested only via automated comparisons with human-annotated benchmarks, with no reported sample size or usability assessment. The intended audience appears to be robotics researchers and designers~\cite{robo-crafter-2025}. Meanwhile, Blox‑Net~\cite{blox-net-2024} uses \ac{VLM} to generate block assembly plans directly from language instructions, enabling end-to-end planning and physical execution with a robot arm. The system was evaluated in simulation using 100 instruction-task pairs but was not tested with users and no usability validation was conducted~\cite{blox-net-2024}. Systems such as Evolution 6.0~\cite{evolution-2025} expand on this by integrating \ac{VLM} and 3D generation to autonomously design tools and action sequences for tasks in novel environments. It was evaluated technically across 30 simulated tasks, but does not include user testing or usability validation. Like the other systems, it is positioned toward autonomous robotic agents, not end-user customization or accessibility~\cite{evolution-2025}. 

Augmented Body Communicator~\cite{genAI-to-design-robots-2025} was co-designed with six participants, which included deaf, blind, and sighted individuals, to create expressive robotic gestures through \ac{LLM} prompting and sketch-based interaction. This study included qualitative feedback and participant reflection, offering insight into usability, though the primary focus remained on gesture design and interaction modalities rather than robot behavior, role, or appearance~\cite{genAI-to-design-robots-2025}. 

A different study found that university students with disabilities used \ac{GenAI} to navigate academic barriers and reported a lack of institutional support and limited avenues for creative agency~\cite{university-students-genAI-2025}. This highlights a broader need for inclusive design frameworks that move beyond task assistance toward empowering disabled users to shape the technologies they interact with. In our study, participants leverage \ac{GenAI} as a tool to boost creativity when creating speculative robot-assisted social dining scenarios (i.e., visual literacy). While \ac{GenAI} has been used to design for \ac{HRI} as seen above, it is still not being used to empower \ac{PwD} to express their ideas. Our research aims to make this gap closer. 

\begin{table*}[t!]
\centering
\caption{Study participant demographics. All participants had experience using Assistive Technology (i.e., technology assisting with motor tasks) and reported dining in public spaces 1--5 times per month.}
\label{tab:participants}
\Description[Study Participant Demographics]{This table summarizes participant demographics, including age, gender, race, self-described disability, required assistance during dining, occupation, chosen scenario for the storyboard, and assistive technology use. It shows that participants varied widely in age, included both female/male, all used some form of assistive technology, and all reported to dine in public spaces approximately 1--5 times per month.}

\begin{tabular}{l l l l l l p{2.8cm} l}
    \toprule
    \textbf{ID} & \textbf{Age} & \textbf{Gender} & \textbf{Race} & \textbf{Self-described Disability} & \textbf{Assistance Type} & \textbf{Occupation} & \textbf{Chosen Scenario}\\
    \midrule
    P1 & 35--44 & Female & White & Quadriplegic & Full Meal & N/A & Family dinner\\
    P2 & 65--74 & Female & White & Osteoarthritis; Shoulder Injury & Cutting/Meal Prep & Respiratory Therapist & Brunch w/ friends\\
    P3 & 55--64 & Female & White & Spinal Muscular Atrophy & Full Meal & Social Worker; ED at Non-Profit & Business meal\\
    P4 & 55--64 & Female & Mixed & Parkinson's/Partial Paralysis & Full Meal & Remote Therapist & Business meal\\
    P5$^\dagger$ & 35--44 & Female & White & Intention Tremors (Hand/Body) & Full Meal & N/A & Brunch w/ friends\\
    P6 & 25--34 & Male & White & Multiple Sclerosis & Cutting/Drinking & Accountant & Business meal\\
    \bottomrule
\end{tabular}

\vspace{0.5em}
\parbox{0.95\linewidth}{$^\dagger$ P5's caregiver was present during the study and assisted them in understanding scenarios. Due to visual impairments, P5 found it difficult to view images generated with \textsc{Speak2Scene}, so we excluded related questions during the exit interview.
}

\end{table*}

\subsection{Speculative Design of Assistive Technologies}
Speculative design is a method that explores possible futures by creating artifacts, scenarios, or systems that provoke reflection and imagination, rather than solving immediate problems~\cite{speculative-2015, speculative-design-2022, co-speculation-uw}. Rooted in critical design~\cite{critical-design}, it challenges assumptions and invites users to question, re-imagine, and re-frame possibilities. Rather than focusing solely on feasibility, speculative design emphasizes desirability, ethics, and broader societal implications~\cite{speculative-2015}.

In the context of assistive technologies, speculative design has been increasingly adopted to move beyond medical (deficit-based) model of disability~\cite{speculation-disability-2022}. Speculative methods have been applied to \ac{HRI}, engaging older adults and aging researchers in envisioning future assistive robots that emphasize autonomy and resilience, rather than framing aging as a problem to be solved~\cite{aging-2018}. The authors of the paper~\cite{aging-2018} highlight that robots designed for successful aging population needs to consider potential disabilities, but should not discount the autonomy and resilience of older adults. The paper employs techniques of collaborative map making~\cite{collab-map}, artifact analysis~\cite{artifact-analysis}, and envision robot possibilities~\cite{envision-robot} as part of their speculative participatory design method~\cite{aging-2018}. 

Other projects have used speculative and fictional approaches to challenge dominant assumptions in HRI and inspire critical engagement on sociotechnical futures~\cite{HRI-speculations-2024,speculative-hri-2025}. For instance, researchers illustrate how bringing art into robotics can create room for speculative \ac{HRI}~\cite{HRI-speculations-2024}. They further talk about co-designing interactions between workers and a speculative baggage handling robot (that can assist with manual labor) at an airport. The study resulted in coming up with meta-behaviors allowing knowledge translation between different parties~\cite{HRI-speculations-2024}. Desjardins et al.~\cite{desjardins} uses co-speculative design to imagine alternative ideas for internet of things in diverse homes. Winkle~\cite{speculative-hri-2025} suggests that speculative design is a legitimate approach in \ac{HRI} since the field already depends heavily on imagining future scenarios.

While these projects do not directly engage disabled users, they illustrate how speculative methods can broaden design imagination and provoke reflection on inclusion and possible futures. Robot-assisted social dining is a relatively new concept, and its future possibilities are underexplored. Speculative methods offer a promising approach to anticipate, question, and shape these futures, which we explore in this paper by engaging in speculations with \ac{PwD}.

\subsection{Assistive Robots for Dining}
Research in robot-assisted feeding has primarily focused making technical improvements to the act of feeding (e.g, bite acquisition~\cite{8624330, transfer-2019, flair-2024, kiri-spoon-2024, generalizing-skewering, adaptive-raf-2020, single-utensil, visual-imitation, sundaresan2022learning}, bite transfer~\cite{feel-the-bite, belkhale-2022, lorenzo-2023}), emphasizing independence and have focused on testing them in-the-lab settings~\cite{ada-2024, active-robot-feeding-2020, Candeias_2018_ECCV_Workshops, song2012novel, 7140110, 8593525, exploring-preferences-2020, fang-2018, eeg-intent, in-lab}. More recently, studies have begun to address social dimensions of robot-assisted feeding, making technical strides and integrating design principles from social dining to show how such systems might function in out-of-the-lab settings~\cite{voicepilot-2024, lessons-learned-2025, Feast-2025}. 

In social dining scenarios, systems must be attuned to physical assistance but also to conversational flow and group behavior. For example, bite-timing models trained on human-human interaction data help robots avoid interrupting social cues during meals~\cite{bite-timing-2023}. Similarly, research on group activity recognition in restaurant settings has shown that understanding shared dining phases enables robots to offer assistance without disrupting the social rhythm of the table~\cite{group-activity-recognition-2022}. These studies were conducted in controlled environments and did not include \ac{PwD}. Conversely, solitary dining systems focus primarily on promoting user independence and reducing caregiver reliance. Real-world evaluations of robot-assisted feeding have shown positive outcomes in terms of safety and comfort, though persistent barriers remain, including cost, complexity, and limited adaptability~\cite{exploring-preferences-2020}. A paper by Nanavati et al.~\cite{lessons-learned-2025} described the deployment of a robot-assisted feeding system in homes, offices, and cafeterias. This work, grounded in participatory design research, demonstrated how sustained in-the-wild trials can increase user independence and control during meals. The deployment aspect of this paper exemplifies how iterative field testing across real-world contexts can reveal the social, technical, and logistical challenges that shape long-term adoption of assistive technologies~\cite{lessons-learned-2025}.

Another paper, VoicePilot~\cite{voicepilot-2024}, demonstrates the feasibility and initial user acceptance of \ac{LLM}-based voice control for robotic feeding. The system was evaluated with 11 older adults, where they used the \ac{LLM}-powered speech interface to control an Obi feeding robot~\cite{obi-robo}. The study included qualitative feedback on usability and expressiveness. However, while the system enabled sequential natural-language instructions, it also introduced challenges such as latency and unintended behaviors from ambiguous commands~\cite{voicepilot-2024}.

FEAST~\cite{Feast-2025} introduces a robot-assisted feeding framework that includes an \ac{LLM}-driven personalization pipeline where users express preferences through speech or text, and the system converts these into structured behavior tree modifications using GPT-4o. FEAST emphasizes robot transparency and allows users to create custom gestures to control the robot. The authors conducted a formative study with 21 \ac{PwD} to understand their needs and performed a 5-day in-home deployment with two community researchers to study meals across real-world scenarios. An occupational therapist unfamiliar with the system also completed a series of personalization tasks, validating the system’s adaptability and safety~\cite{Feast-2025}. 

Compared to VoicePilot~\cite{voicepilot-2024}, which focused primarily on sequential voice commands, FEAST~\cite{Feast-2025} offers a broader, more interactive personalization framework. Nanavati et al.~\cite{lessons-learned-2025} presented lessons learned from deployment of the robot in several out-of-home and out-of-lab contexts, while FEAST~\cite{Feast-2025} focuses on personalization to users and deployments in-home settings. Taken together, while each system advances specific aspects of robot-assisted feeding, a unified solution that supports both socially embedded use and individual customization, especially across diverse users and contexts, remains an open challenge.

\section{Method}
Robot-assisted social dining has the potential to support \ac{PwD} experience fewer missed social and professional connections. However, there remains a gap in understanding the needs of such systems in out-of-home and out-of-lab settings with diverse users. To address this, we employed a speculative participatory design method combined with semi-structured interviews to capture participants’ perspectives. Our study was approved by the University of Michigan's IRB\#HUM00273373.

\subsection{Participants}
A total of $6$ participants with disabilities participated in our study. A full description of our sample can be found in Table~\ref{tab:participants}. Upon completion of the design session, participants received a compensation of \$50. Our inclusion criteria consisted of any \ac{PwD} who needed any level of assistance (from either a caregiver\footnote{A caregiver is someone who assists another person in helping accomplish Activities of Daily Living (e.g., eating, dressing).} or a family member/friend) to eat. Recruitment was conducted through purposive outreach via disability advocacy networks, direct contact with care organizations, and snowball sampling~\cite{snowball-2019}, which is a method where participants are recruited by leveraging the network effect of each participant and organization. 

Given the speculative and exploratory nature of our research, we adopted the principle of information power~\cite{Sample-size-2016} to stop recruitment. This method emphasizes the relevance, richness, and specificity of a sample over its size. In line with critiques of the saturation concept in reflexive thematic analysis~\cite{Saturate-2019}, our aim was not statistical generalizability or theme exhaustion. Instead, we worked towards conceptual sufficiency, enabling the development of rich, nuanced themes within and across participant accounts. 

\begin{figure}[t!]
    \centering
    \includegraphics[width=\linewidth]{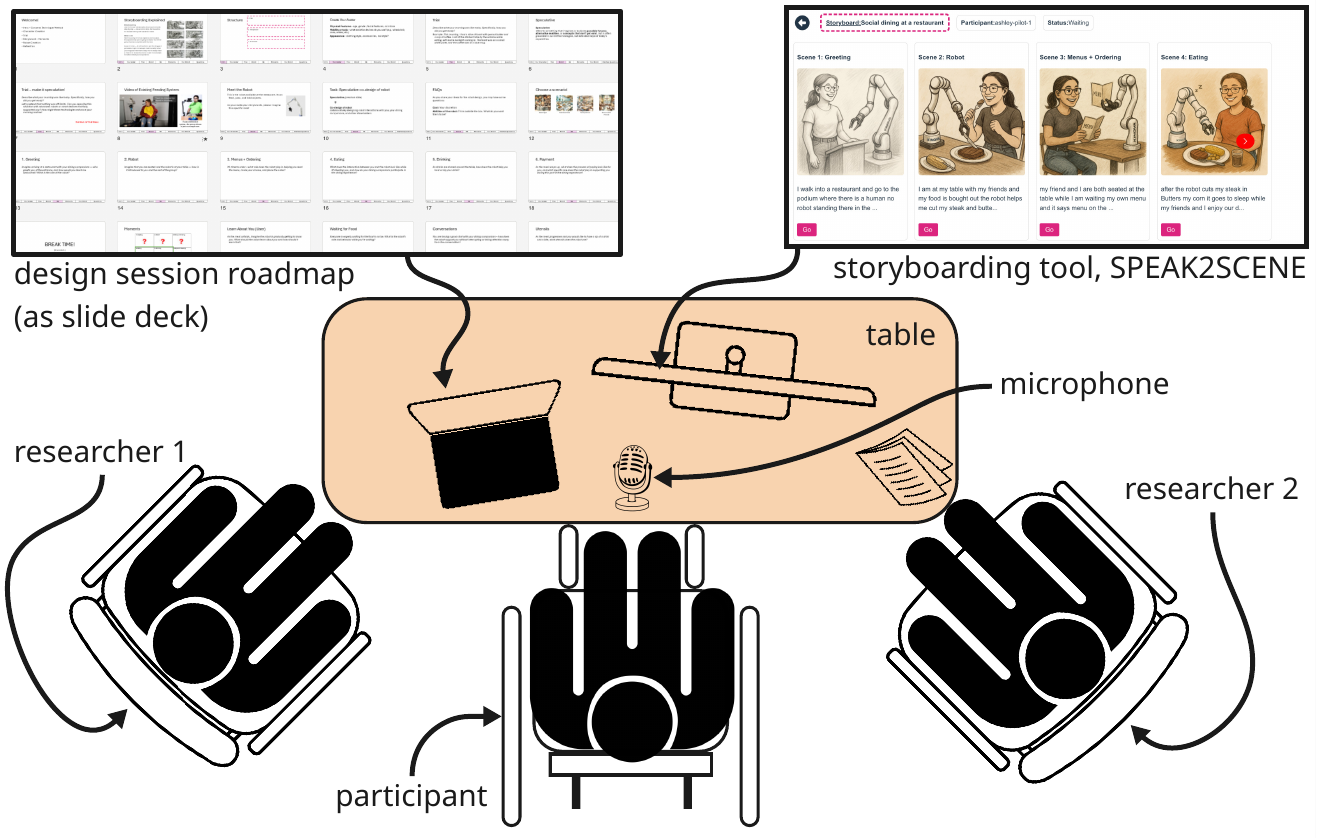}
    \caption{Study Setup. The study was conducted either at the university or at the participant's home with the same set-up.}
    \Description[Study Setup]{This illustration is a graphic portraying the study setup. There is a table with a monitor displaying Speak2Scene, the AI-based storyboarding tool, and a second computer displaying the slides. There is a microphone to capture participants' ideas. There are two additional seats for the two researchers to sit and be involved in the study.}
    \label{fig:study-setup}
\end{figure}

\subsection{Materials}
The study employed a multi-component setup (See Figure~\ref{fig:study-setup}) that included: (1)~\textsc{Speak2Scene}, a custom storyboarding tool powered by a \ac{VLM}, and (2)~design session roadmap (slide deck) to support participants during the session.
 
\subsubsection{\textsc{Speak2Scene}}
\label{sec:speak2scene}
Design research shows that storytelling methods can help participants envision new futures and express their experiences more creatively~\cite{storytelling-for-novel-methods}. Storyboarding extends storytelling by giving participants a structured, visual format for illustrating experiences and drawing out design ideas~\cite{SB-in-PD-2016, lit-review-storyboarding, storyboarding-in-automotive, storyboarding-power-equip}. However, because storyboarding typically involves hand-sketching, it may not be accessible for our target group. So, we built a customized voice-based storyboarding tool using ReactJS~\cite{ReactJS} web framework for the front-end, and the Firebase~\cite{Firebase} platform for the database and to deploy the tool. A video of the tool being used is provided in the supplemental materials. Participants used \textsc{Speak2Scene} to generate speculative images for various scenes of social dining. To generate an image, participants interacted with the tool using their own voice (i.e., natural language) to indicate the type of image they wished to create. Their speech was captured by the laptop's microphone and was converted to text using a React Speech Recognition hook~\cite{react-speech-recognition}. To generate the image, participants' speech, in addition to prompt engineering detailed in Appendix~\ref{appendix:prompt-engin-speak2scene}, is sent to OpenAI's \texttt{gpt-image-1}~\cite{gpt-image-1} \ac{VLM} through an API call. After approximately 30-45 seconds, an image based on the participants' prompt is generated and displayed on the computer monitor. If the participant is not satisfied with the image, they had the option to try again. 

\subsubsection{Design Session Roadmap}
A researcher-led slide deck was created to serve as the study roadmap. It helped orient participants in the session and introduce key concepts such as robot-assisted feeding, storyboarding, and example prompts (e.g., robot gestures, group settings, interaction styles). The full slide deck is provided in the supplemental materials.

\begin{figure*}[t]
    \centering
    \includegraphics[width=\textwidth]{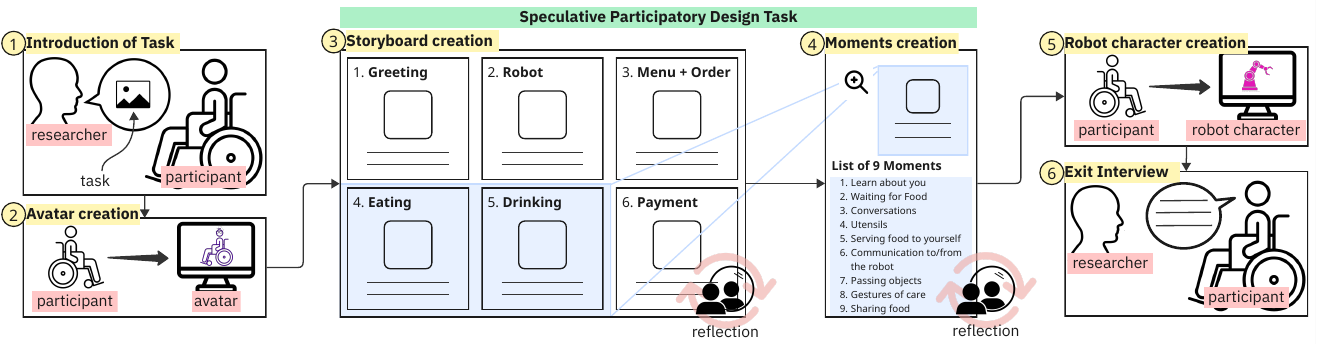}
    \caption{Procedure of the design study included: (1)~introducing the task to the participant, (2)~creating an avatar based on their lived experience to be represented in the task, (3)~creating a storyboard of a generic experience of dining at a restaurant, (4)~zooming into 9 key moments (indicated in blue), (5)~creating an imagined robot character, and (6)~exit interviews to supplement participant's design process. During steps (3) and (4), participants self-reflected to supplement their images.}
    \Description[Procedure for design study]{The figure is a flow diagram titled “Speculative Design Task”. It illustrates six main steps in a participatory design activity, using icons, labels, and arrows to show the sequence. The steps are as follows: (1) Introduction of Task/Scenario – a researcher introduces the task or scenario to a participant (who has a disability); (2) Avatar Creation – the participant creates an avatar or character on a computer; (3) Storyboard Creation – six storyboard panels are created (which include, Greeting, Robot, Menu + Order, Eating, Drinking, and Payment. Each panel includes space for an image and text. Participants reflect after creating an image for each panel; (4) Moments Creation – a participant chooses from a List of 9 Moments (Learn about you, Waiting for Food, Conversations, Utensils, Serving food to yourself, Communication from the robot, Passing objects, Gestures of care, and Sharing food). Each moment has its own template card with space for an image and description; (5) Robot Character Creation – the participant designs a robot character (represented by an icon of a robotic arm on a screen); (6) Exit Interview – the researcher and participant engage in a final interview and reflection. Arrows connect the steps in order, showing progression from introduction, avatar and storyboard creation, moment detailing, robot character design, and ending with the exit interview.}
    \label{fig:procedurefig}
\end{figure*}

\subsection{Procedure}
The procedure is summarized in Figure~\ref{fig:procedurefig} with details about researcher roles and study sessions below.

\begin{figure*}[t!]
  \centering

  \begin{subfigure}[b]{0.48\linewidth}
    \centering
    \includegraphics[width=\textwidth]{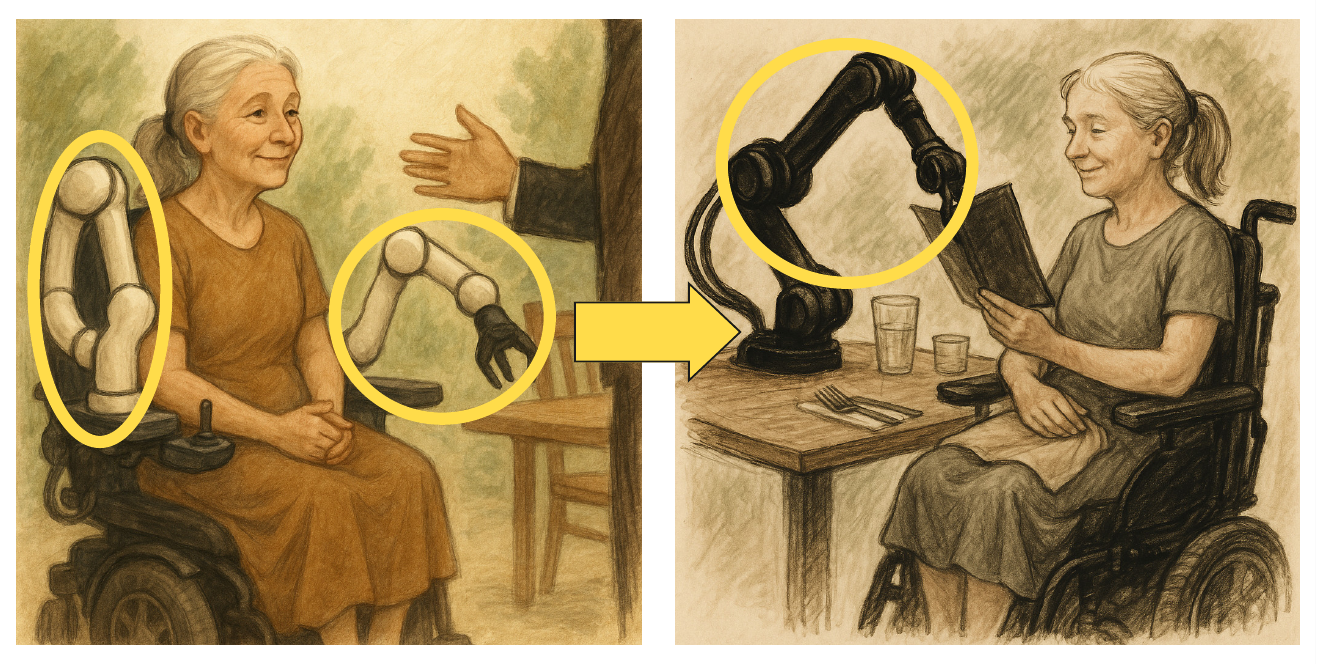}
    \caption{Robot color~(P3). (Left) \textit{``I arrive at the restaurant... We're greeted by the person...I'm not really seeing the role of the robot at this point...unless it's to maybe pull out a chair or move things around so that I can get to the table''} (Right) \textit{``We are seated at the table. I pull my wheelchair up to the edge of the table and maneuver the robot which is black not white to move silverware and drinkware to where I need to be able to reach it...picking up the menu to look at the menu...happen right there putting my napkin on my lap''}}
    \Description[Iteration of robot color]{Images present the iteration of robot color.}
    \label{fig:new-sub1}
  \end{subfigure}
  \hfill
  \begin{subfigure}[b]{0.48\linewidth}
    \centering
    \includegraphics[width=\textwidth]{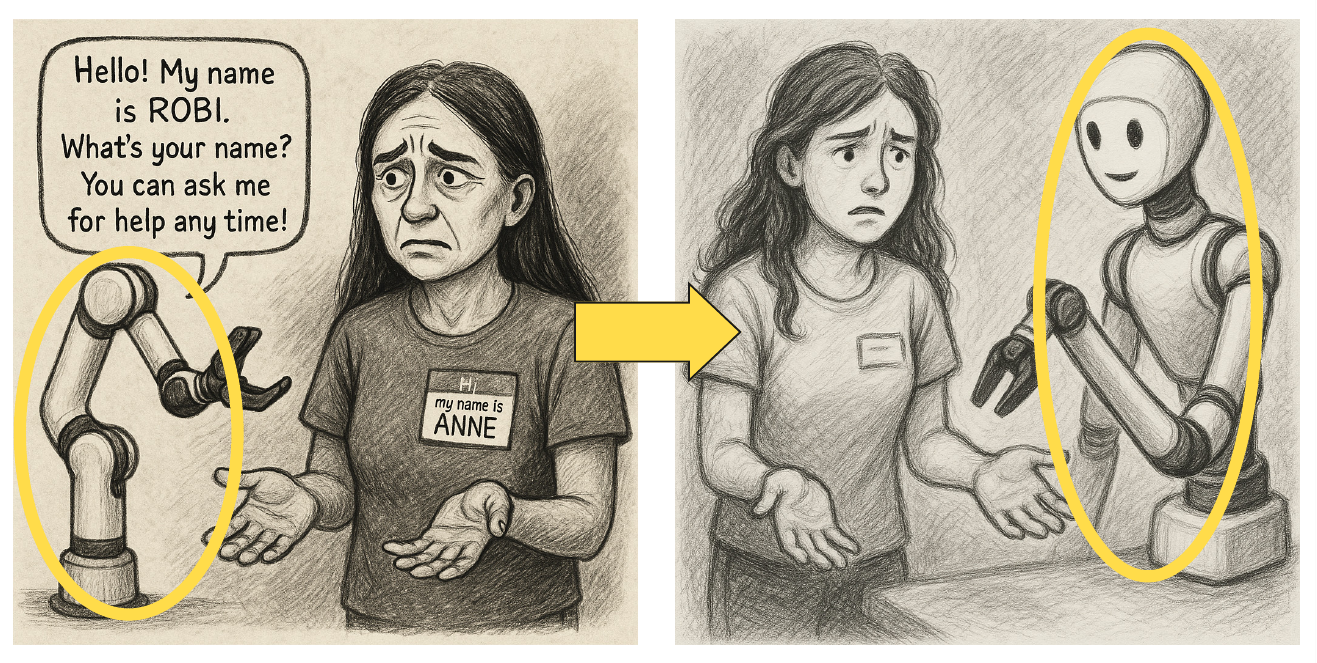}
    \caption{Robot Personality~(P5). (Left) \textit{``I would want the robot to introduce itself with a name and ask me what my name is and tell me how to ask it for help if I need help''} (Right) \textit{``[We] went to [restaurant] and we had called ahead to let them know that we would need a robot to help me when we got to the restaurant...and they told us that the robot, Claire, would be helping us. So Claire came out to a table and said, `Hello! Who will I be assisting today?' and I told Claire, `Hi! I'm [name]. I will need your help today...'''}}
    \Description[Iteration of robot personality]{Images present the iteration of robot personality}
    \label{fig:new-sub2}
  \end{subfigure}

  \vspace{1em}

  \begin{subfigure}[b]{\linewidth}
    \centering
    \includegraphics[width=0.60\textwidth]{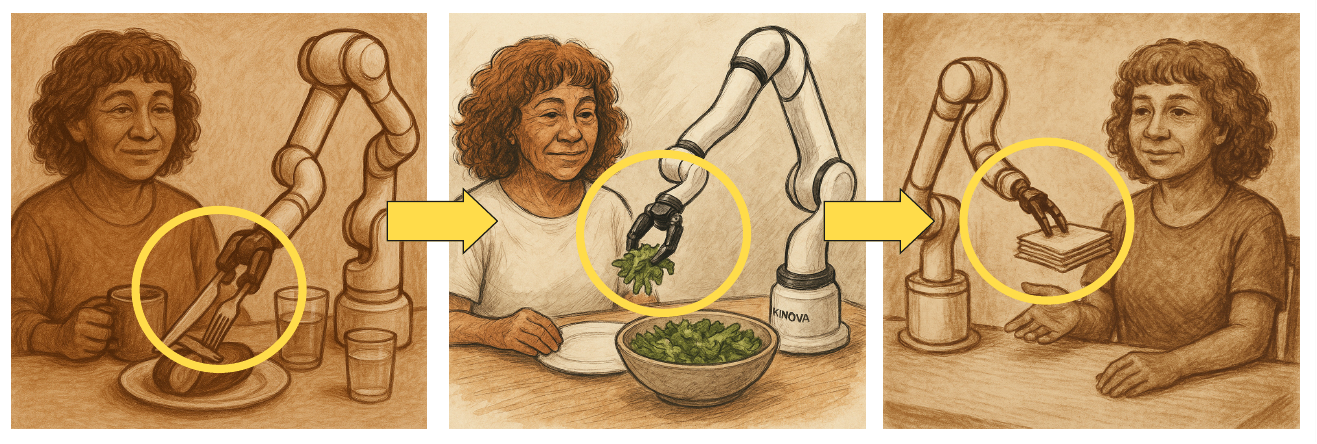} 
    \caption{Robot Role~(P4). (Left) \textit{``I think it's important for the robot to be adaptive at using all different types of utensils including the ability to hold a knife and a fork at the same time to cut meat or a tough vegetable like a eggplant...''} (Middle) \textit{``if there's a shared dish in the middle...and I would like more of it,...robot [would] be signaled by myself then be able to reach into the middle of the table for the shared dish and put some on my plate which is the correct amount that I want...I might say I want more salad and...the robot would be intelligent enough to choose that specific dish in the middle of the table and give me that exact amount on my plate...''} (Right) \textit{``...for example the disabled person could command the robot pick up napkins on my left side and hand it to the person sitting directly across the table''}}
    \Description[Iteration of robot role]{Images present the iteration of robot role in regards to using different types of utensils.}
    \label{fig:new-sub3}
  \end{subfigure}

  \caption{Iteration of robot during speculative design. The caption captures the prompt by participant to create scenes.}
  \label{fig:iteration-examples}
\end{figure*}

\subsubsection{Researcher Roles}
All sessions were led by two of the authors of this paper. One author served as the ``facilitator'' to ease participant interactions and supervised the storyboarding tool. Another author served as the ``interviewer'', taking notes and following up with clarifying questions. Both researchers maintained a reflexive stance throughout the process, taking field notes and engaging in regular debriefings to document emerging impressions, adjust facilitation strategies, and reflect on their influence on the design process. The interdisciplinary research team included expertise in human-centered robotics, disability studies, and qualitative research, which informed both the study setup and subsequent analysis.

\subsubsection{Design Sessions}
\label{sec:study-session}
Design sessions were conducted in-person, either at the participants' homes or at the university, depending on participant preference and accessibility needs. Each session was conducted individually and lasted approximately 75–90 minutes. These sessions followed a predefined protocol with flexibility to be adjusted to dive deeper into participants' areas of interest within the study. At the start of each session, participants were welcomed by the facilitator and provided with an overview of the study. The facilitator reviewed the informed consent form, addressed any questions, and obtained verbal recorded consent.

The sessions began with a brief orientation using a facilitator-led slide deck introducing key concepts such as robot-assisted feeding,  speculative design, and Kinova arm~\cite{kinovaGen3} (the robot we expect to use for social dining in the future stage of this research). An image of the Kinova arm was provided to ground participants’ speculations. Then, participants were introduced to the storyboarding tool, \textsc{Speak2Scene}. They were shown examples of prompts and system-generated scenes to familiarize themselves with the tool. Following this, participants were asked to create an avatar that reflected their identity, which served as the ``main character'' of the storyboarding. After, they engaged in a trial session where they used \textsc{Speak2Scene} to create a few trial images to oriented with the tool. 

Once participants were oriented with the tool, they engaged in a speculative design task using \textsc{Speak2Scene}. Figure~\ref{fig:iteration-examples} shows examples of design iteration as created by participants. Participants were supported by the facilitator, who navigated the tool. Each participant was first encouraged to generate a full six-panel storyboard representing a social dining scenario of their choice (e.g., family dinner, brunch w/ friends, business meal, date night). They verbally described imagined scenes involving themselves, their dining companions, and the robot in the chosen context. After generating the complete storyboard, participants were asked to elaborate on nine ``key moments'' within their scenes. These included specific interactions like sharing food or waiting for food. For each selected moment, the facilitator posed reflective questions to prompt participants to clarify what the robot was doing, why it behaved that way, and how it could act more appropriately in that context. The reason behind diving deeper into key moments of a social dining experience was to understand the nuances in expected robot role and behavior in diverse intricate aspects of social dining, a very specific perspective not considered in prior work.

Throughout the design process, the facilitator asked semi-structured follow-up questions to clarify participants' intentions, encourage elaboration, and explore deeper meanings behind their scene choices. These questions were designed to elicit insight into preferences for robot behavior, social presence, communication, and adaptability. Clarifications were prompted both proactively (based on themes of interest) and responsively (based on participant scenes).

After engaging in the storyboarding session, participants were asked to create an image for the robot character where they described how they envision the robot's appearance. Following the storyboarding experience, participants were invited to reflect on details about the robot, as by then they had developed enough familiarity with the idea of a robot assisting them in social dining. This enabled them to imagine robot's additional role and appearance, which may not have been possible at the outset. At the end, participants were invited to reflect on their experience, share any closing thoughts, and optionally keep a digital copy of their storyboard.

\section{Analysis}
\label{sec:analysis}
We conducted reflexive thematic analysis on the interview transcripts following the framework outlined by Braun and Clarke~\cite{Braun-Clarke-2006}, which includes familiarization, initial coding, theme development, review, definition, and final write-up. Half of the interview transcripts were coded by one researcher and the other half by a second researcher. First, both researchers individually conducted open coding on the transcript from the 6 design sessions using MAXQDA~\cite{maxqda2025} to identify key elements such as robot behaviors, communication modes, and contextual cues.

The coding captured both semantic features (e.g., what actions the robot performs) and latent features (e.g., underlying expectations about autonomy, dignity, and social comfort in different dining contexts). Once all the transcripts were coded, both researchers had discussions for 16+ hours to form a consensus on the codes and categorized them based on the research questions. Disagreements during discussions were handled via collaborative reflexive thematic analysis. Finally, themes were generated from these codes. Codes and themes are visualized in Appendix~\ref{appendix:figs-for-acc}. These themes highlight key areas where future efforts should focus to meet user needs in robot-assisted social dining and they are: 
\begin{itemize}
    \item \textbf{Interaction Ecology in Robot-Assisted Social Dining} (RQ1, Section~\ref{RQ1-results})
    \item \textbf{Context-Sensitive Robot Behavior in Social Dining} (RQ2, Section~\ref{RQ2-results})
    \item \textbf{Robot Role During and Outside of Mealtime} (RQ3, Section~\ref{RQ3-results})
    \item \textbf{Perceived User relationship with Robot} (applicable to RQ1, RQ2, and RQ3, Section~\ref{allRQ-results})
\end{itemize}

While there are studies that analyze visual storyboards~\cite{storyboarding-2022, storyboards-2016, mybodymyskeleton-2025, moodboards-2015}, analyzing the images in our study would not be appropriate. Participants used \ac{GenAI} to create storyboards in our study and while \ac{VLM}s are good at capturing ideas, they can hallucinate~\cite{llm-hallucinations, hallucinations-image-llm-1, hallucinations-2024-2}, thereby introducing aspects not explicitly mentioned by participants or disregarding others that it cannot fully capture. It is critical that researchers do not overuse or over-rely on \ac{GenAI} in design. To account for this, in our study, the storyboards were treated as a tool for participant visual literacy, which refers to the capacity to treat the images as starting points, rather than the end result, to encourage speculations about the robot.   

\section{Results}
Our results include insights about the interaction ecology in robot-assisted social dining, context-sensitive robot behavior in social dining, the role of the robot during and outside mealtime, and the perceived user relationship with the robot.

\subsection{Interaction Ecology in Robot-Assisted Social Dining}
\label{RQ1-results}
\begin{figure*}[t!]
    \centering
    \includegraphics[width=\textwidth]{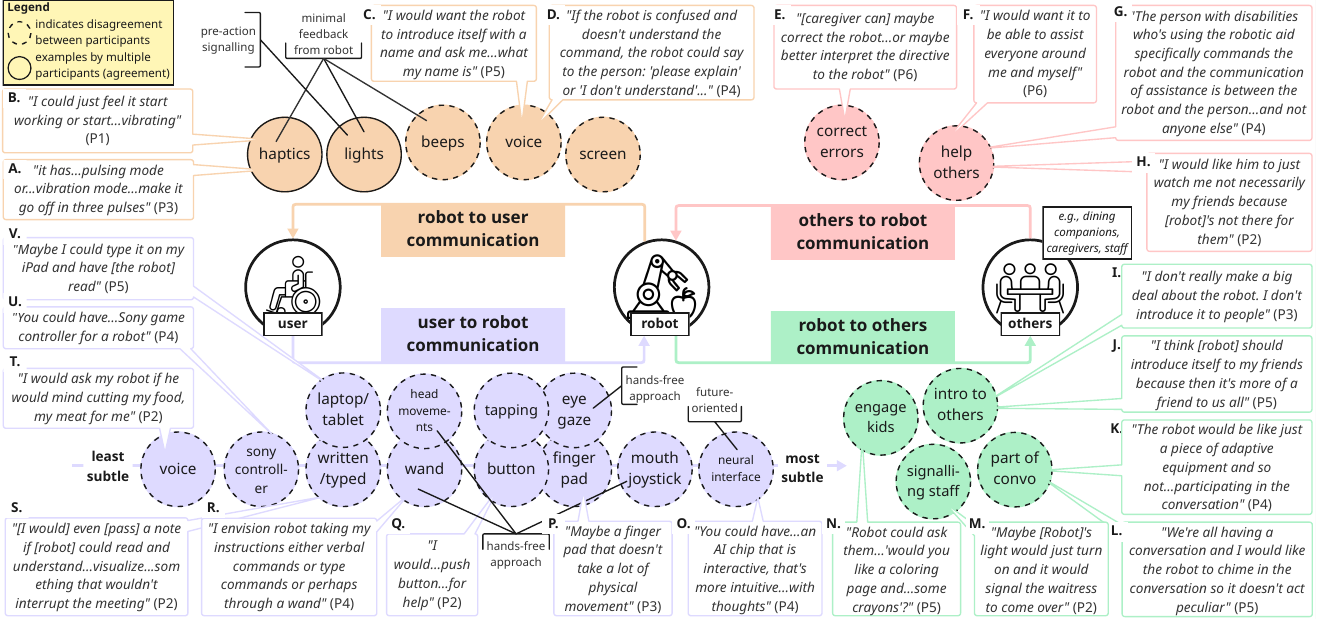}
    \caption{Participants' viewpoints on the Interaction Ecology in Robot-Assisted Social Dining described in Section~\ref{RQ1-results}. Further details can be seen in Table~\ref{tab:participant-coms},~\ref{tab:rq1-acc-table-a}, and~\ref{tab:rq1-acc-table-b} in Appendix~\ref{appendix:figs-for-acc}}
    \label{fig:RQ1-results-fig}
    \Description[Communication pathways in robot-assisted dining.]{The figure maps communication pathways among the user, the robot, and others (such as dining companions, caregivers, or staff). Four categories of interaction are shown with colored clusters and example participant quotes and they are as follows: (1) Robot-to-user communication (orange): via lights, haptics, beeps, voice, or screen, including minimal feedback or pre-action signaling; (2) User-to-robot communication (purple): input modalities ranging from least to most subtle, including voice, laptop/tablet, written/typed text, wand, buttons, tapping, head movements, eye gaze, finger pad, mouth joystick, and neural interfaces; (3) Others-to-robot communication (pink): caregivers may help correct robot errors or assist others; (4) Robot-to-others communication (green): engaging kids, introducing itself to others, signaling staff, or participating in conversation. Each pathway is illustrated with clustered keywords/codes and direct quotes from participants, highlighting preferences, disagreements, and envisioned interaction designs.}
\end{figure*}

Our thematic analysis revealed that participants envisioned interaction with the robot in terms of both \textbf{user-initiated commands} and \textbf{robot-initiated feedback}, when interacting with caregivers and dining companions. Refer to Figure~\ref{fig:RQ1-results-fig} for a summary of results.

\subsubsection{Adaptable and Multimodal User-to-Robot Communication}
Participants described a wide spectrum of modalities for giving commands to the robot, from low-effort tactile methods (e.g., button press, finger pad, joystick, wand) to hands-free approaches (e.g., eye gaze, head movement, voice commands). Some emphasized the value of having options that fit their changing needs:
\begin{displayquote}
\textit{``I could press buttons [if] I'm feeling good...[If] I'm tired, I might need help choosing or [if] my hand is sore, [there] could have a little section of things I might need more help with.''}~(P2) 
\end{displayquote}
Others saw potential in voice interaction, for instance, one participant suggested, \textit{``I could ask the robot like what is gluten-free on the menu?''}~(P5), envisioning the robot not only for assistive feeding, but also as an accessible source of information during dining. This also raised practical considerations, as participants stressed that such voice features would need to adapt to noisy settings:
\begin{displayquote}
\textit{``And perhaps voice recognition so it drowns out other people's voices during the dining experience and ambient noise''}~(P4)
\end{displayquote}
Still, others speculated about future-oriented modalities, imagining \ac{AI}-based or even neural interfaces:
\begin{displayquote}
\textit{``I think the best case scenario would be that it would have...an AI interface...maybe they even have an electrode transmitting thoughts. And the robot is intuitive so when the person in the wheelchair thinks, ‘hand me my coffee,’ it does it.''}~(P4)
\end{displayquote}
This framing emphasized flexibility: participants desired multiple pathways for interaction to accommodate changing abilities, fatigue, or situational constraints.

These findings echo the design principles presented in Nanavati et al.~\cite{design-principles-2023}, where participants valued subtle modalities in the robot system, such as a button press, for their reliability and low intrusiveness, and raised concerns about voice commands disrupting social interactions. However, our speculative approach extended beyond concrete bite-initiation options to envision a broader ecology of multimodal interaction, including the robot being \textbf{adaptive and predictive}. This highlights a design frontier not addressed in Nanavati et al.~\cite{design-principles-2023}: rather than optimizing one or two modalities, future feeding systems may need to flexibly support a suite of input options that users and/or the robot can switch between depending on environment, mood, or physical needs.

\subsubsection{Minimal and Unobtrusive Robot-to-User Communication}
In contrast to expansive command options, participants emphasized that robot feedback should remain minimal and unobtrusive. One participant envisioned, \textit{``just beep''}~(P6), illustrating how simple cues could convey status without disrupting social flow. Beeps, lights, or simple screen displays were preferred over verbose speech, which risked burdening conversations. Importantly, participants highlighted the need for \textbf{pre-action signaling} (e.g., a light or sound before the robot moves), framing it as a trust-building mechanism that prevents surprise and supports predictability. As one participant explained, 
\begin{displayquote}
\textit{“Or maybe if there's a red light like, a stop—like red, yellow, and green—so I know what it’s about to do.”}~(P4)
\end{displayquote}
Such anticipatory cues were seen not just as functional indicators but as social reassurance, helping participants feel prepared for the robot’s actions. 

In Nanavati's et al.'s work that explored design principles~\cite{design-principles-2023}, the authors discussed preferences for the robot resting poses and unobtrusive bite transfer and placed less emphasis on proactive robot signaling. Our participants’ focus on anticipatory cues suggests additional layers of trust-related design in robot-assisted feeding that have not yet been uncovered. Whereas findings from Nanavati et al.~\cite{design-principles-2023} were centered on avoiding error and ensuring subtlety of the robot system, our results stress the social reassurance that comes from transparency-users want to know not only \textit{what} the robot is doing but also \textit{when} it is about to act. For the successful deployment of robot systems, the complementary approach brought to light needs to be considered.

\subsubsection{Communications Involving Others}
Participants also extended interaction beyond the dyad of user-and-robot. Some envisioned \textbf{caregiver-robot teaming}, where caregivers could intervene or redirect the system when needed. This highlights that participants did not envision the robot as a replacement for caregivers, but rather as a collaborator working alongside them. As one participant described,
\begin{displayquote}
\textit{``I think the person who’s disabled could directly give commands to the robot, but if there’s an aide there and the robot makes a mistake, the aide could correct the robot or better interpret what the person meant, because the aide has a better understanding and history of that person’s wants and needs.''}~(P4)
\end{displayquote}
Others considered the robot’s role in relation to dining companions. Here, perspectives diverged: some wanted the robot to remain invisible in the social exchange, while others suggested that it should introduce itself or politely acknowledge companions, positioning the robot as a subtle but socially aware presence at the table. For example, one participant imagined, 
\begin{displayquote}
\textit{``Maybe it could have a prerecorded little blurb—like a taped message—you press one or two, and it says: ‘Hi, this is [robot’s name].’''}~(P2) \end{displayquote}
Yet others envisioned the robot to be relational, engaging and contributing to conversations with the user and dining companions. This extends the work presented by Nanavati et al.~\cite{design-principles-2023}, where participants emphasized caregiver–robot teamwork but did not explicitly consider how robots should manage their social presence with dining companions. Our speculative framing moves interaction design into the realm of \textbf{group dynamics}. The notion that a robot might acknowledge, converse with, or otherwise integrate into shared dining interactions broadens the design space from feeding mechanics to \textbf{social awareness}. This highlights a unique contribution of our study: situating assistive feeding robots not only as functional tools but also as actors embedded within the rituals of group dining.

\subsection{Context-Sensitive Robot Behavior in Social Dining}
\label{RQ2-results}
\begin{figure*}[t!]
    \centering
    \includegraphics[width=\textwidth]{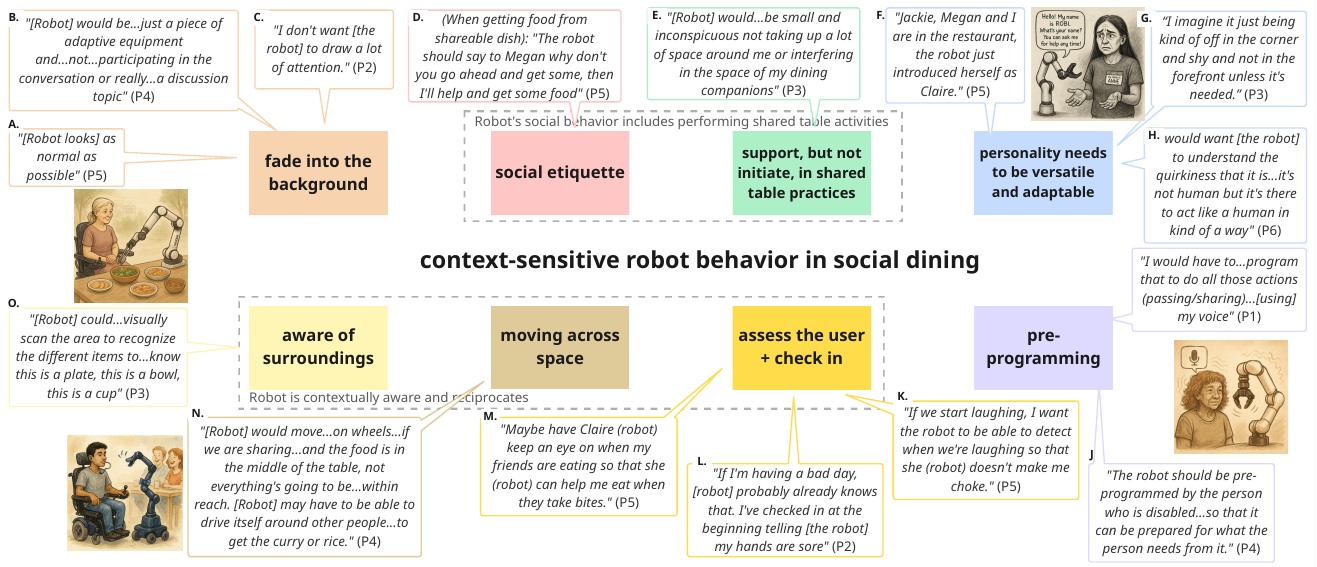}
    \caption{Participants' perspective on context-sensitive robot behavior in social dining was generally in agreement. The four images within this figure were created by the participants using \textsc{Speak2Scene}, where each image captures some ideas. Further details can be seen in Appendix~\ref{appendix:figs-for-acc}.}
    \label{fig:RQ2-results-fig}
    \Description[Participant feedback on context-sensitive robot behavior in social dining]{This illustration is an image summarizing some of participants perspectives. They are grouped into various sub-themes that are mentioned in Section~\ref{RQ2-results}. The subthemes are: (1) fade into background, (2) robot's social behavior includes performing shared table activities, (3) personality needs to be versatile and adaptable, (4) robot is contextually aware and reciprocates, and (5) pre-programming. In this diagram, some of these subthemes are further divided and there are specific participant quotes that are supporting them.}
\end{figure*}

This theme uncovered how participants view the robot's behavior. \textit{Behavior} is defined as the way in which one \textit{conducts} oneself to others~\cite{dict-behavior}. In a broad sense, participants agree that they want the robot to \textbf{fade into the background}, something that has also been found in the study of Nanavati et al.~\cite{design-principles-2023}. However, regarding the broad robot behavior and robot movements, participants envision the robot being \textit{``gentle''} and \textit{``delicate''}~(P6). Further, according to design canvases presented by Axelsson et al.~\cite{canvases}, in the context of robots, specifically social robots, the idea of \textit{robot behavior} can be defined under four tenets, which include \textit{personality}, \textit{social behavior}, \textit{contextual awareness}, and \textit{personalization}. Our findings are grouped into sub-themes inspired by these groupings and are described below. It is also summarized in Figure~\ref{fig:RQ2-results-fig}.

\subsubsection{Robot's Personality needs Adaptation}
According to Merriam-Webster dictionary~\cite{dict-personality}, \textit{personality} is defined as the \textit{characteristics} of an individual. 
Participants generally presented a wide spectrum of opinions on whether they see the robot as having a personality or not. As a participant detailed in one of the scenes,
\begin{displayquote}
\textit{``I would say thank you very much to my [robot] for helping and giving me a really nice evening out''}~(P2), portraying a bond, where the user expresses gratitude to the robot.
\end{displayquote}
A few other participants directly mentioned that they don't see the robot having a personality. Participants also see the robot having \textit{funny} or \textit{quirky} personality without having features of a person. For instance, one participant mentioned,
\begin{displayquote}
\textit{``I would want it to be that kind of silly... does not compute kind of mentality... I don't want it to be person-like, because that would irritate me a bit''}~(P6)
\end{displayquote}

Other participants mentioned that they would treat the robot as another individual with a name and gender at the table. There isn't a clear consensus as to what participants prefer when it comes to the robot's personality, but it is clear that the robot's \textbf{personality needs to be versatile and adaptable} based on the user. 

\subsubsection{Robot's Social Behavior Includes Performing Shared Table Practices}
By \textit{shared table practices}, we refer to coordinated social actions that occur around the dining table, encompassing both the passing of items (e.g., salt/pepper, napkins) and sharing food (e.g., serving food from a common plate, splitting portions). These practices are often both functional and social. Our results show that participants expected the robot to \textbf{support, but not initiate, in shared table practices}. For example, a participant described that the \ac{PwD}
\begin{displayquote}
\textit{``using the robot aid could instruct the robot through whatever means to lift a glass to do the toast''}~(P4)
\end{displayquote}
As seen in this example, the nuance in this idea of the robot helping in these shared activities but not engaging in them is that participants envision the robot as an extension of themselves. This suggests that they expect to have full ``agency'' to control the robot, a preference that seems to emerge as they see the robot as an extension of themselves and their body, enabling them to perform actions that they currently are unable to do. Further, participants expect the robot to be intelligent enough to be able to \textit{``lift a plate or reach across or maybe move around a person to get the curry''}~(P4). Participants want to feel independent in all realms of social dining and not just in the aspect of feeding. They also see the robot understanding the \textbf{social etiquette} to serve socially appropriate amounts of food onto one's own plate. 

\begin{figure*}[t]
    \centering
    \includegraphics[width=\textwidth]{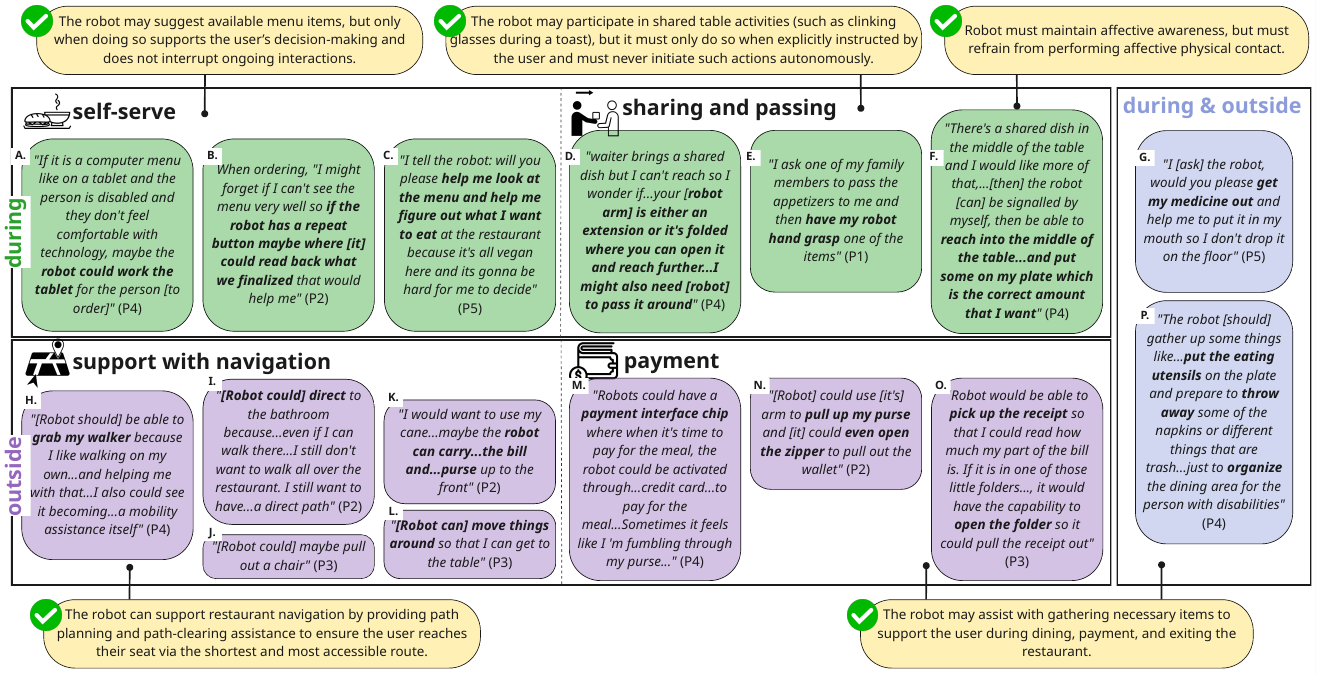}
    \Description[Participants' standpoints on robot role]{This illustration summarizes participants' perspectives on robot role. They are grouped into three categories: (1) during mealtime, (2) outside of mealtime, and (3) during \& outside of mealtime. Under the first category (during), there are two subthemes: (a) self-serve and (b) sharing and passing. Under the second category (outside), there are two subthemes: (a) support with navigation and (b) payment. Under the third category (during \& outside), there is a miscellaneous subtheme. Each subtheme has several supporting participant quotes.}
    \caption{Participants' standpoints on robot role. Top row shows their perspectives on robot role \sethlcolor{customgreen!40}\hl{during mealtime} and bottom row shows their perspectives \sethlcolor{custompurple!40}\hl{outside of mealtime}. The right-most section shows the robot requirements that are applicable for \sethlcolor{customblue2!40}\hl{during and outside} of mealtime. Rules synthesizing ``dos'' and ``don'ts'' are captured in the yellow boxes on the borders. See Appendix~\ref{appendix:figs-for-acc} for additional details.}
    \label{fig:RQ3-results-fig}
\end{figure*}

\subsubsection{Robot is Contextually Aware and Reciprocates}
\label{sec:context}
Participants described that the robot needs to be well \textbf{aware of surroundings} and needs to understand the layout of the table setup effectively. They express the need to ensure that the robot can semantically understand the items on the table and their presence relative to the robot arm. For example, a participant explained that,
\begin{displayquote}
\textit{``if you have... soup and sandwich on the table at the same time, it's got to know where the bowl is versus where the sandwich on the plate is''}~(P3)
\end{displayquote}

Context is not limited to the food items or tableware, but also may include the type of dining experience. For instance, participants picturing dining in a buffet-style restaurant described that the robot should:
\begin{displayquote}
\textit{``be able to move across space and be mobile, which will help the person with disabilities obtain food or drink that they want in the dining scenario''}~(P4)
\end{displayquote}
Thus, participants not only saw the robot moving in place to help with feeding and performing tasks on the table, but they also envisioned the robot \textbf{moving across space}, perhaps on wheels. 

Participants expressed intent through multiple modes -- a few are explicit and many are subtle. One participant explained that his dining companion(s) can sometimes:
\begin{displayquote}
\textit{``just tell when [he is] more fatigued... because [he is] not as excited in the conversations... or maybe [his] hand is getting a little more shaky''}~(P6)
\end{displayquote}
Similarly, participants expected the robot also to be able to \textbf{assess the user and periodically}, yet not frequently, \textbf{check-in} with the user. 

\subsubsection{User-based Robot Learning is Essential}
Participants emphasized the importance of the robot learning about their needs, and there was consensus that, in addition to adapting throughout the meal, the robot should also include an element of \textbf{pre-programming}, allowing the user to manually ``show'' it some of their preferences. One participant suggested incorporating a \textit{``teaching mode''} within which they are able to \textit{``teach the robot''} using multi-modal interaction~(P3). When describing what the robot should learn, participants do not envision the robot learning anything about their disabilities, but rather just understanding their needs and preferences. As one participant mentioned,
\begin{displayquote}
\textit{``I think that the robot shouldn't know...our disabilities''}~(P5), and another participant described that the robot \textit{``should learn how big to cut things so they are not too big...or too small''}~(P3)
\end{displayquote}
Further, participants also suggested that the robot should learn their voice, especially if the interaction involved user voice input, to help separate their communication with the robot from that of other dining companions.

\subsection{Robot Role During and Outside of Mealtime Assistance}
\label{RQ3-results}
As defined by the Merriam-Webster dictionary~\cite{dict-role}, \textit{role} is a \textit{function} performed. And, in the context of this theme, we explore the robot's \textit{function} during and outside of mealtime. Authors of a Monolith~\cite{Armada2023}, a piece of literature that is developed to help train people in restaurants/hospitality industries, mention the idea of `restaurant sequence of service'. Inspired by this idea, we defined `during mealtime' to be stages where the diner would be placing an order, waiting for food, and eating/drinking. And, we defined `outside of mealtime' to be stages where the diner would generally not be interacting with dishes or drinks. The results on the robot role are summarized in Figure~\ref{fig:RQ3-results-fig}.

\subsubsection{During Mealtime}
Participants described their needs from the robot during mealtime to be a wide spectrum, including needing robot help to \textbf{self-serve}, which can inherently make them feel more independent. An aspect of being able to serve oneself is the ability to order a dish independently, which entails the need to have access to a menu and place an order. A role that some participants see the robot playing is in helping them \textbf{place orders}, especially if the order placement procedure (i.e., reading a menu or using a tablet) is inaccessible to users due to their disability. As one participant described,
\begin{displayquote}
\textit{``the menus are on the table and sometimes hard to reach or they are slippery and they might fall...or someone's having problems turning it over because there's...food items on the front and the back that the robot could turn it over''}~(P4)
\end{displayquote}

Furthermore, some participants mentioned the inaccessibility of a restaurant buffet and explained roles that the robot could play to help them feel independent and included. For instance, one of the participants mentioned:
\begin{displayquote}
\textit{``I'd like to be able to use my cane and have [robot] hold plates for me''}~(P2)
\end{displayquote}

Echoing the popular adage, ``sharing is caring'', several participants expected that the robot would also be able to help with \textbf{sharing and passing} of dishes, silverware, and other items during mealtime. One participant expressed that when her dining companion wants to try some food on her plate, she would \textit{``instruct the robot to...pass [her] plate of food to the person so they can get some of it''}~(P4). The role of the robot could be to serve as an extension to the user's arm where, for instance, the robot \textit{``could reach [the shared dish] for me and bring it closer''}~(P2). 

\subsubsection{Outside of Mealtime}
Participants outlined their broad needs from the robot outside of mealtime. While most participants agreed that they wished to be greeted by a human wait-staff with minimal social interaction with the robot, there was a variety of other expectations from the robot when \ac{PwD} entered/exited the restaurant with their dining companions. 

Some participants expected \textbf{support with navigation}, which can come in different forms for different people. As one participant mentioned,
\begin{displayquote}
\textit{``when the meal wraps up, how can I safely get out?''}~(P6).
\end{displayquote}
This suggests that the participant wants the robot to not only anticipate an end to the dining experience, but also create a navigation plan to safely support them in exiting the restaurant. Additionally, participants see the robot helping them with carrying miscellaneous items (e.g., purse, carry-out box), which can be difficult for them to do if they are navigating a restaurant environment while also being disabled. In general, \ac{PwD} prefer to plan their navigation routes ahead of time (especially if they use wheelchairs or walkers), and they expect the robot to play a role in helping them with that. As one participant mentioned, the robot could:
\begin{displayquote}
\textit{``pull out a chair or...move things around so that I can get to the table''}~(P3)
\end{displayquote}

Participants outlined the robot's role to also support with \textbf{payment}. More broadly, \ac{PwD} might struggle with being able to retrieve their mode of payment either from their wallet or their purse, as decreased mobility may make this process burdensome. As one participant expressed,
\begin{displayquote}
\textit{``I would get my wallet out and...I can't exactly get my card out so...I could have the robot delicately be able to grab my card out and...pay at the table''}~(P6)
\end{displayquote}

Another participant described a social component of the robot's role by illustrating an example of her arguing with her dining companions over the person picking up the check. Humorously, the participant suggested that the \textit{``robot chimes in and tells who gets to pay [and] the robot chooses me''}~(P5).

\subsection{Perceived User Relationship with the Robot}
\label{allRQ-results}
\subsubsection{Human Agency in Robot Control}
Participants mentioned they would prefer to have high-level control of the robot, making them feel independent. Further, they felt that having too much automation can be restrictive, even in instances of social dining, so striking a balance between automation and control is important.

\subsubsection{Adaptability of Robot Appearance}
\begin{figure}[t!]
    \centering
    \includegraphics[width=\linewidth]{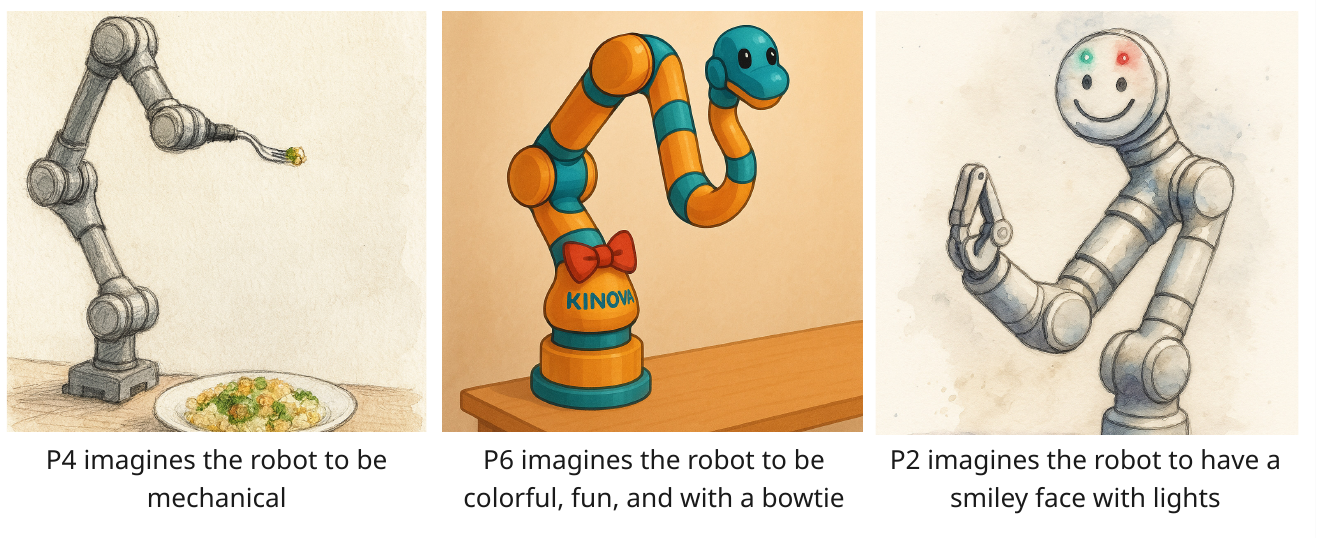}
    \caption{Speculative robot ideas by participants.}
    \label{fig:appearance-results-fig}
    \Description[Example speculative robot ideas generated by participants]{The figure presents three illustrations of how some participants imagined the robot’s appearance: (1) Left (P4): A mechanical looking robot arm. And the caption reads, ``P3 imagines the robot to be mechanical.'' (2) Middle (P6): A bright, colorful robot arm is shown in orange and teal stripes. It has cartoon-like eyes, a playful snake-like shape, and wears a red bowtie. And the caption reads, ``P6 imagines the robot to be colorful, fun, and with a bowtie.'' (3) Right (P2): A gray robotic arm with a round head features a smiley face. Two indicator lights—one red and one green—are above the eyes. And, the caption reads, ``P2 imagines the robot to have a smiley face with lights.''
    Together, the images highlight how some participants envisioned robot appearances differently, ranging from utilitarian and discreet to playful and expressive.}
\end{figure}

According to the Merriam-Webster dictionary~\cite{dict-appearance}, \textit{appearance} is defined as the outward \textit{look}. As the saying goes, ``beauty is in the eye of the beholder'', and while all participants agreed on the idea of the robot blending in, they held \textbf{diverse} and sometimes \textbf{contrasting views on robot embodiment, size, and color} (see Figure~\ref{fig:appearance-results-fig}).

Some participants preferred for the robot to embody and act like a human, including having \textit{``pretend food''} on the robot's plate so \textit{``that it looks like the robot is really eating''}~(P5). One of the reasons for preferring this was that the robot would not stand out when among bystanders. Others preferred the robot to be a \textit{``basic robot''} or a \textit{``mechanical thing''}~(P2). There were also participants who felt that the robot should have a non-human-like face, perhaps a screen, that could be a \textit{``circle like...a smiley face...could have...eyes''}~(P2). 

Many users had opinions on the color and size of the robot. Because some users preferred the robot to simply blend in and not stand out, they wanted the robot to be \textit{``black and not white''}~(P3). As one participant mentioned,
\begin{displayquote}
\textit{``I want it to be small...so that it wouldn't distract''}~(P1).
\end{displayquote}
While some prefer the robot to have more neutral colors, some participants suggested that the robot should have ``fun'' colors. As described by a participant,
\begin{displayquote}
\textit{``I would not want it to look the stereotypical white/black robot...I kind of want it to have some color...make it look a little fun...have a bow tie}~(P6)
\end{displayquote}
This shows that preferences for the robot’s appearance varied, with some participants desiring playful or colorful designs, while others favored more neutral and subtle appearances. 

\subsubsection{Who Owns the Robot?}
Participants had varying ideas regarding the robot's ownership. Many participants considered the cost of the robot as a potential decision point. Some considered the robot to be a medical device, thereby seeing potential for insurance to cover the costs and enabling them to own the robot as a personal device. Others also mentioned hiring the robot \textit{``just for the...outing''}~(P2). A few others also saw the potential of restaurants owning them such that there is \textit{``one arm per table''}~(P6), thereby enhancing consistency of robot presence not just for \ac{PwD}, but rather for any customer at the restaurant. Some participants pointed out issues with the cleaning process, especially when the device is shared across diners~(P3), highlighting the need to design robot-assisted social dining systems with hygiene at the forefront. 

\section{Discussion}
\label{discussion}
While prior work, such as Nanavati et al.~\cite{design-principles-2023} and formative study in FEAST~\cite{Feast-2025}, identified the importance of user control, minimal communication, and unobtrusive assistance, these studies primarily focused on the dyad of robot and primary user. Our speculative \ac{PD} approach extends this by examining the triad of primary user, robot, and dining companions, uncovering forms of social coordination that occur in shared dining environments.

Prior work emphasized the process of feeding only. However, our participants highlighted the importance of shared activities such as passing dishes, coordinating toasts, or supporting social rituals. This resonates with hospitality practices~\cite{Armada2023}, where service quality is evaluated holistically. Participants additionally envisioned roles beyond feeding, including suggesting menu items, assisting with path- planning or clearing, and preparing for transitions such as paying or exiting the restaurant. These findings point towards the need for a mobile manipulator capable of supporting a suite of dining-related tasks, not solely feeding.

Prior work has also not extensively examined robot appearance or personality. Our study reveals that these elements are not aesthetic details but meaningful contributors to comfort, trust, and social fit. This is especially relevant since there is diversity in our participants' disabilities and social environments. Supporting interchangeable physical features and multiple personality profiles may therefore be essential for future systems.

Finally, our study considers questions about robot ownership and access. Participants expressed differing preferences for owning, renting, or borrowing assistive robots for dining experiences. While some assistive robots (e.g., Kinova~\cite{kinovaGen3}) are covered by insurance, participants noted that access remains limited. Broader availability through insurance, short-term rental models, or restaurant-provided robots could improve accessibility and reduce financial burden on individuals. We now discuss Design Implications stemming from this work that move forward the domain of socially assistive robots for feeding.

\subsection{Design Implications for Robot-Assisted Social Dining as a \textit{White Glove Service}}
Our findings highlight that participants envision robot-assisted feeding not only as a functional interaction where the robot supports feeding, but as a service embedded within the broader experience of social dining -- which we frame using the metaphor of \textit{white glove service}~\cite{linkedin-wgs}, a hospitality concept that emphasizes: (1)~anticipatory assistance, (2)~discretion \& privacy, (3)~attention to detail, (4)~personalization, and (5)~seamless problem resolution. Applying this metaphor to assistive robotics helps articulate what ``good service'' means in highly social, interdependent dining contexts and extends prior work by shifting attention from the mechanics of feeding to the quality of being served.

\subsubsection{Anticipatory Assistance}
Participants preferred robots that could anticipate upcoming needs based on situational cues rather than repeated user commands. Anticipation did not imply full autonomy; instead, it involved identifying opportune moments for support. For example, preparing payment assistance as meal concludes or planning an accessible route when the user is ready to exit. Such timing was interpreted as socially aware and competent, particularly in socially dense settings where excessive prompting could be disruptive.

\colorlet{shadecolor}{gray!95}
\setlength{\FrameRule}{2pt}
\setlength{\FrameSep}{6pt}

\begin{leftbar}
\vspace{0.9\baselineskip}
\begin{minipage}{0.85\linewidth}
\noindent\textbf{Design Implication:} Robot-assisted social dining system should incorporate models for mealtime progression and social context to proactively, but not intrusively, initiate helpful actions.
\end{minipage}
\vspace{0.9\baselineskip}
\end{leftbar}

\subsubsection{Discretion \& Privacy}
Participants emphasized the importance of robots that provide attentive yet unobtrusive assistance. Desired interactions involved periodic check-ins that do not interrupt conversations and a general ability for the robot to ``blend into the background''. Further, participants were not comfortable with robots inferring diagnostic information about their disability; instead preferred the robot to respond to observable symptoms of their disability or explicitly user-provided cues.

\colorlet{shadecolor}{gray!95}
\setlength{\FrameRule}{2pt}
\setlength{\FrameSep}{6pt}

\begin{leftbar}
\vspace{0.9\baselineskip}
\begin{minipage}{0.85\linewidth}
\noindent\textbf{Design Implication:} Robots should employ lightweight and low-interruption communication strategies and prioritize symptom-level sensing and privacy-preserving data practices.
\end{minipage}
\vspace{0.9\baselineskip}
\end{leftbar}

\subsubsection{Attention to Detail}
Participants envisioned robots that are socially aware and capable of adapting to nuances of dining etiquette. This included serving appropriate portion sizes, recognizing turn-taking around shared dishes, and adjusting behaviors based on the social dynamics at the table (e.g., dining with work colleagues vs. with family). Such attention to detail places the robot as a socially competent, rather than a mere appliance.

\colorlet{shadecolor}{gray!95}
\setlength{\FrameRule}{2pt}
\setlength{\FrameSep}{6pt}

\begin{leftbar}
\vspace{0.9\baselineskip}
\begin{minipage}{0.85\linewidth}
\noindent\textbf{Design Implication:} Robots should integrate models of dining etiquette and tabletop social dynamics enabling moment-to-moment adaptation during shared activities.
\end{minipage}
\vspace{0.9\baselineskip}
\end{leftbar}

\subsubsection{Personalization}
Personalization emerged as a valued dimension and it spanned interaction preferences, robot embodiment, and personality. Diversity in participant disability shaped varying expectations for robot form, size, visual appearance, and manner of interaction. Preferences for embodiment or personality were influenced not only by accessibility needs but also by users' social identities and comfort in public dining settings.

\colorlet{shadecolor}{gray!95}
\setlength{\FrameRule}{2pt}
\setlength{\FrameSep}{6pt}

\begin{leftbar}
\vspace{0.9\baselineskip}
\begin{minipage}{0.85\linewidth}
\noindent\textbf{Design Implication:} Robot-assisted social dining should support personalization across both software and hardware, such as adjustable appearance, adaptable interaction styles, and configurable personalities.
\end{minipage}
\vspace{0.9\baselineskip}
\end{leftbar}

\subsubsection{Seamless Problem Resolution}
Participants expect robots to avoid mistakes where possible but to recover from errors without causing additional social burden. Many described scenarios where caregivers could team up with the robot to intervene as intermediaries, enabling quick resolution while maintaining the user's dignity and social flow.

\colorlet{shadecolor}{gray!95}
\setlength{\FrameRule}{2pt}
\setlength{\FrameSep}{6pt}

\begin{leftbar}
\vspace{0.9\baselineskip}
\begin{minipage}{0.85\linewidth}
\noindent\textbf{Design Implication:} Robot-assisted social dining systems should incorporate mechanisms for rapid handover to caregivers and support collaborations within the triad of user, robot, and caregiver.personalities.
\end{minipage}
\vspace{0.9\baselineskip}
\end{leftbar}

\subsection{Reflection on Speculative Participatory Design Method using GenAI}
Our Design Implications were facilitated by the usage of speculative \ac{PD} with \ac{GenAI}. There were both benefits and shortcomings to using \ac{GenAI} as a tool for speculative \ac{PD}. Some participants found value in \textsc{Speak2Scene} because it made them feel hopeful and helped them think outside the box about how they envision the robot system. However, on the flip-side, the tool centered on instances where the generated images slightly misrepresented participants’ abilities or assistive needs. While participants generally understood these limitations, hallucinated images could occasionally distract from the design conversation, though at times they also served as prompts that encouraged further creative thinking. Participant fatigue also appeared to play a role. Early in the study, participants were more motivated to engage in trial-and-error with the image generation process, but as the session progressed, they became less inclined to re-try generation attempts when mistakes occurred.

\section{Limitations \& Future Work}
In this paper, our design sessions lasted 75-90 minutes, with some of them even extending beyond due to the needs of participants. While such sessions were extremely informative and provided rich insights, they can feel long despite breaks during the study. Future studies should aim to decrease the time spent. Additionally, as mentioned in Section~\ref{sec:analysis}, we predominantly used the storyboards created using \textsc{Speak2Scene} for visual literacy, which refers to the capacity to treat the images as starting points for speculation. Exploring avenues to analyze visual storyboards with images created using \ac{GenAI} would be an interesting future approach. Moreover, our participants were all based in the United States and were primarily white (5 white/1 mixed) and female (5 female/1 male). Future work should expand to more diverse participant groups to ensure findings generalize across varied cultural and demographic contexts. Our participants were also only \ac{PwD}, who are the primary users of the system, but engaging secondary users (e.g., caregivers, restaurant staff) would further add depth to our paper's conclusion and could provide complementary perspectives on how robot systems can adapt to broader social dining needs.

With the stated limitations, our findings remain crucial to the assistive robotics research field. Our main insight encourages designing robots for social dining that embody the principles of \textit{white glove service} -- supporting multimodal inputs and unobtrusive outputs, exhibiting contextually sensitive social behavior that prioritizes the user, expanding roles beyond feeding, and adapting to relationships at the dining table. These insights provide a foundation for future robot-assisted social dining systems to not only support the act of eating but the full experience of social dining.

\begin{acks}
GPT-4o helped with the grammar of this paper. All ideas and thoughts are original to the authors. Icons from the NounProject (https://thenounproject.com/) were used in Figures~\ref{fig:study-setup}, \ref{fig:procedurefig}, \ref{fig:RQ1-results-fig}, \ref{fig:RQ3-results-fig}: "Seat Passenger" by daniel solis sanchez; "Chair Top View", "Computer Top View" by Romain Bernard; "Laptop" by Nick Taras; "Microphone" by Smashing Stocks; "ask pictures" by corpus delicti; "Wheelchair Accessibility User" by Suncheli Project; "reflection" by Tom Ingebretsen; "monitor" by Heztasia; "robotic arm" by Bernd Lakenbrink; "Meeting" by Good Wife; "Navigation" by Eskak; "Food" by Milky - Digital innovation; "exchange" by Hayashi Fumihiro; "robot arm" by Icon Queen. We thank our participants for sharing their time, perspectives, and experiences, as well as their caregivers for their support in making this research possible. We also thank the organizers of several disability organizations who invited us to attend events and engage with the community. We thank Dr. Long-Jing Hsu, Connor Williams, Francesca Cocchella, Grace Pan, and all members of Robot Studio for piloting our study and providing feedback on our paper. We thank Dr. Laura Murphy for help with piloting and Prof. Leia Stirling for help with IRB. 
\end{acks}

\bibliographystyle{ACM-Reference-Format}
\bibliography{biblio}

\appendix
\section*{Appendices}
\section{Prompt Engineering for \textsc{Speak2Scene}}
\label{appendix:prompt-engin-speak2scene}
In addition to capturing participants' ideas, we engineered the prompts to include two features: (1)~participant's avatar, and (2)~an image of Kinova Gen 3~\cite{kinovaGen3}. Information about the necessities of these features are explained in Section~\ref{sec:study-session}. The prompt we engineered is included below:

\subsection{Prompt to generate an avatar:}
\begin{quotation}
\texttt{Generate a photorealistic image based on the following description: <self-description by participant>}
\end{quotation}

And, the generated photorealistic image based on self-description by the participant is converted to a sketch/brush-based illustration using the following prompt:

\begin{quotation}
\texttt{Use sketchy, brush-based illustration techniques, like a concept to generate an avatar or character for the following prompt: <prompt>}
\end{quotation}

The \texttt{<prompt>} will be a text-based description of the photorealistic image created of the participant's self-description. The prompt to convert the photorealistic image to text is the following: 
\begin{quotation}
\texttt{Based on the provided image, generate a detailed and objective description of the character. Include physical attributes, clothing and appearance, any assistive or mobility features the character may use, and notable facial expressions or body language that might hint at their personality or emotional state. Do not attempt to identify or name the character—focus solely on descriptive analysis.}
\end{quotation}

\subsection{Prompt to generate scenes for the storyboard:}
Description of the Kinova~\cite{kinovaGen3} is provided as the following text:
\begin{quotation}
\texttt{The robot is a Kinova robotic arm that is lightweight, assistive manipulator designed for close human interaction. It features six or seven degrees of freedom with smooth and articulated joints, allowing for versatile object manipulation. Commonly used in assistive technology, it can be mounted on wide variety of platforms (e.g., wheelchair, table, tripod stand, etc.) to help users with tasks like eating, grabbing objects, or performing personal care. Its safe, low-force design and compatibility with various control interfaces (e.g., joystick, sip-and-puff, or voice commands) make it ideal for individuals with limited mobility.}
\end{quotation}

An image of the robot arm is also provided which is then converted into text-based description. 
\begin{quotation}
\texttt{Based on the provided robot image, provide a brief description of the emobodiment of the robot. Capture all the details such that I can replicate the image without ever looking at the image.}    
\end{quotation}

To generate the scene, we provide the participant's self-description, participant's description based on the photorealistic image, robot's physical description, and robot's image description from above. This gets combined with the new prompt/scene description that the participant provides.

\begin{figure*}[t!]
  \centering

  \begin{subfigure}[h]{\linewidth}
    \centering
    \includegraphics[width=\textwidth]{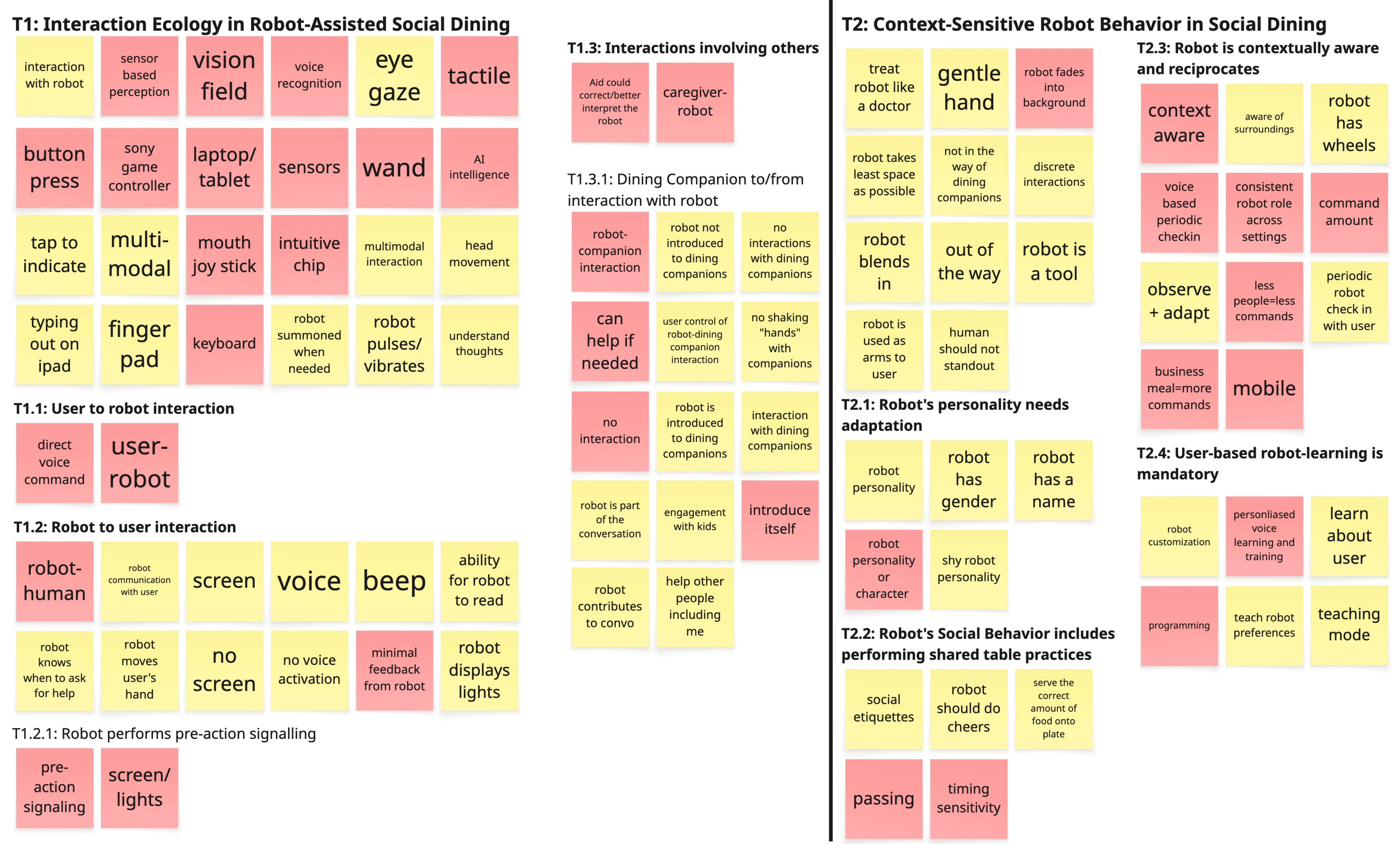}
    \caption{Thematic Analysis: Generation of themes 1 \& 2 from codes.}
    \Description[Codes for Thematic Analysis and generation of Themes 1/2]{Codes for Thematic Analysis and generation of Themes 1/2}
    \label{fig:new-sub11}
  \end{subfigure}

  \vspace{1em}

  \begin{subfigure}[h]{0.9\linewidth}
    \centering
    \includegraphics[width=\textwidth]{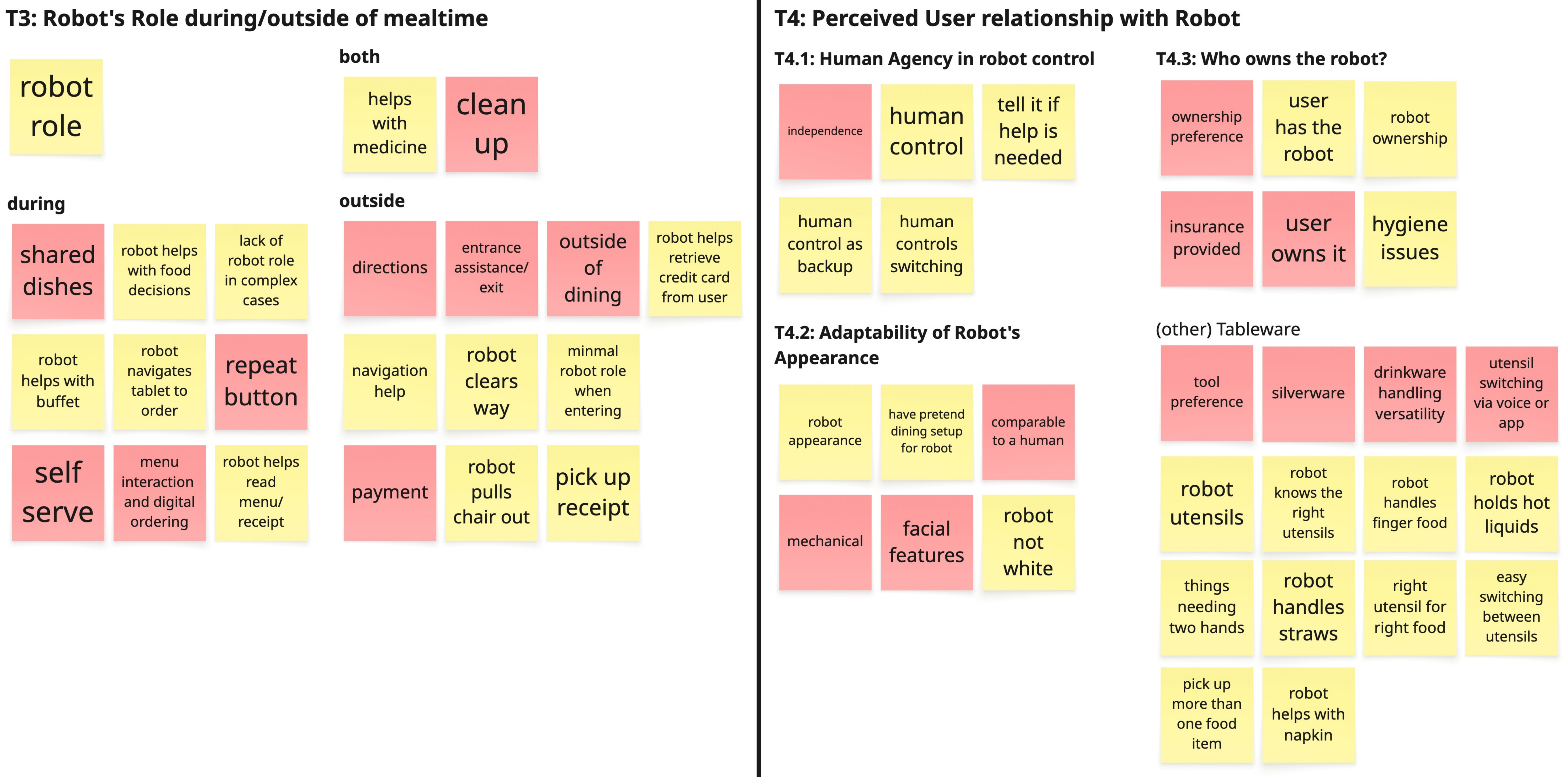} 
    \caption{Thematic Analysis: Generation of themes 3 \& 4 from codes.}
    \Description[Codes for Thematic Analysis and generation of Themes 3/4]{Codes for Thematic Analysis and generation of Themes 3/4}
    \label{fig:new-sub33}
  \end{subfigure}

  \caption{Thematic Analysis Codes}
  \label{fig:thematic-analysis}
\end{figure*}

\section{Additional Figures and Tables}
\label{appendix:figs-for-acc}
\begin{itemize}
    \item Thematic Analysis: Clustering of Codes into Themes in Figure~\ref{fig:thematic-analysis}
    \item Figure~\ref{fig:RQ1-results-fig} summarized in the form of what each participant prefers in Table~\ref{tab:participant-coms}
    \item Figure~\ref{fig:RQ1-results-fig} summarized in Table~\ref{tab:rq1-acc-table-a} and~\ref{tab:rq1-acc-table-b}
    \item Figure~\ref{fig:RQ2-results-fig} summarized as Table~\ref{tab:participants-acc-rq2} 
    \item Figure~\ref{fig:RQ3-results-fig} summarized as Table~\ref{tab:participants-acc-role-rq3}
\end{itemize}

\section{Demographics Questionnaire for Participants}
\label{appendix:questionnaires}

\begin{enumerate}
    \item What is your age range? 
    \item What is your gender identity? 
    \item What is your race/ethnicity?
    \item Do you have a professional occupation? (e.g., Professional Caregiver, Software Engineer, College Student)
    \item Do you have prior experience with assistive technologies?
    \item How comfortable are you with using technology? (e.g., smartphone, computer)
    \item Do you identify as a person with disability that affects your upper limbs?
    \item Describe the nature of your disability.
    \item Do you need assistance while eating?
    \item How do you receive assistance when eating?
    \item How often do you dine in public spaces? (e.g., Restaurants)
    \item How often do you dine alone and with others?
\end{enumerate}

\begin{table*}[t]
\caption{Participant responses for user/robot/others communications. ``\cmark'' indicates that the participant prefers that modality, ``\xmark'' indicates that the participant does not prefer that modality, and ``\umark'' indicates either the participant did not have an opinion about the modality or was not brought up during the study.} 
\Description[Participant responses for user/robot/others communications]{This table captures the participant preferences for if they envision robot-to-user, user-to-robot, robot-to-others, and others-to-robot communications to take place.}
\label{tab:participant-coms}

\centering
\setlength{\tabcolsep}{5pt}
\renewcommand{\arraystretch}{1.15}

\begin{tabular}{c
cccccc
cccccccccccc
cccc
cc}
\toprule
\multirow{2}{*}{\textbf{P. ID}}
& \multicolumn{5}{c}{\textbf{R $\rightarrow$ U}}
& \multicolumn{12}{c}{\textbf{U $\rightarrow$ R}}
& \multicolumn{4}{c}{\textbf{R $\rightarrow$ O}}
& \multicolumn{2}{c}{\textbf{O $\rightarrow$ R}} \\
\cmidrule(lr){2-6}
\cmidrule(lr){7-18}
\cmidrule(lr){19-22}
\cmidrule(lr){23-24}

& \rotatebox{90}{lights}
& \rotatebox{90}{haptics}
& \rotatebox{90}{beeps}
& \rotatebox{90}{voice}
& \rotatebox{90}{screen}

& \rotatebox{90}{neural}
& \rotatebox{90}{mouth joy.}
& \rotatebox{90}{finger pad}
& \rotatebox{90}{tapping}
& \rotatebox{90}{eye gaze}
& \rotatebox{90}{buttons}
& \rotatebox{90}{head move.}
& \rotatebox{90}{wand}
& \rotatebox{90}{tablet}
& \rotatebox{90}{written}
& \rotatebox{90}{controller}
& \rotatebox{90}{voice}

& \rotatebox{90}{engage kids}
& \rotatebox{90}{intro others}
& \rotatebox{90}{signal staff}
& \rotatebox{90}{part of convo}

& \rotatebox{90}{correct}
& \rotatebox{90}{help} \\
\midrule

P1 & \cmark & \cmark & \umark & \umark & \umark 
   & \xmark & \xmark & \xmark & \xmark & \xmark & \xmark 
   & \xmark & \xmark & \xmark & \xmark & \xmark 
   & \cmark & \xmark & \xmark & \umark & \xmark 
   & \umark & \xmark \\

P2 & \cmark & \umark & \umark & \cmark & \umark 
   & \umark & \umark & \umark & \umark & \umark & \cmark 
   & \umark & \umark & \umark & \cmark & \umark 
   & \cmark & \xmark & \cmark & \cmark & \xmark 
   & \umark & \xmark \\

P3 & \cmark & \cmark & \xmark & \xmark & \xmark 
   & \cmark & \cmark & \cmark & \xmark & \cmark & \umark 
   & \cmark & \umark & \cmark & \xmark & \xmark 
   & \xmark & \xmark & \xmark & \xmark & \xmark 
   & \xmark & \xmark \\

P4 & \cmark & \umark & \umark & \cmark & \cmark 
   & \cmark & \cmark & \umark & \umark & \umark & \umark 
   & \umark & \cmark & \cmark & \umark & \cmark 
   & \cmark & \xmark & \xmark & \umark & \xmark 
   & \cmark & \xmark \\

P5 & \umark & \umark & \umark & \cmark & \cmark 
   & \xmark & \umark & \umark & \umark & \umark & \umark 
   & \umark & \umark & \cmark & \cmark & \umark 
   & \cmark & \cmark & \cmark & \umark & \cmark 
   & \umark & \cmark \\

P6 & \cmark & \umark & \cmark & \cmark & \umark 
   & \cmark & \umark & \umark & \cmark & \umark & \cmark 
   & \umark & \umark & \cmark & \umark & \umark 
   & \cmark & \umark & \umark & \umark & \xmark 
   & \cmark & \cmark \\
\bottomrule
\end{tabular}
\end{table*}

\begin{table*}[h!]
\centering
\caption{Participants' perspectives on interaction ecology in robot-assisted social dining (Part 1). ``R'' indicates robot, ``U'' indicates user, and ``O'' indicates others (dining companions, staff, caregiver). ``\cmark'' indicates agreement and ``\xmark'' indicates disagreement.}
\label{tab:rq1-acc-table-a}
\Description{Participant quotes reflecting perspectives on interaction ecology in robot-assisted social dining.}

\begin{tabular}{l l l l p{1cm} p{9cm}}
\hline
\textbf{Lbl.} & \textbf{P. ID} & \textbf{Sub-theme} & \textbf{Task} & \textbf{Agree?} & \textbf{Quote} \\
\hline
A & P3 & R $\rightarrow$ U & haptics & \cmark & \textit{"it has...pulsing mode or...vibration mode...make it go off in three pulses"} \\
B & P1 & R $\rightarrow$ U & haptics & \cmark & \textit{"I could just feel it start working or start...vibrating"} \\
C & P4 & R $\rightarrow$ U & voice & \xmark & \textit{"If the robot is confused and doesn't understand the command, the robot could say to the person: `please explain' or `I don't understand'..."} \\
D & P5 & R $\rightarrow$ U & voice & \xmark & \textit{"I would want the robot to introduce itself with a name and ask me...what my name is"} \\
E & P6 & O $\rightarrow$ R & correct errors & \xmark & \textit{"[caregiver can] maybe correct the robot...or maybe better interpret the directive to the robot"} \\
F & P6 & O $\rightarrow$ R & help others & \xmark & \textit{"I would want it to be able to assist everyone around me and myself"} \\
G & P4 & O $\rightarrow$ R & help others & \xmark & \textit{"The person with disabilities who's using the robotic aid specifically commands the robot and the communication of assistance is between the robot and the person...and not anyone else"} \\
H & P2 & O $\rightarrow$ R & help others & \xmark & \textit{"I would like him to just watch me not necessarily my friends because [robot]'s not there for them"} \\
I & P3 & R $\rightarrow$ O & intro to others & \xmark & \textit{"I don't really make a big deal about the robot. I don't introduce it to people"} \\
J & P5 & R $\rightarrow$ O & intro to others & \xmark & \textit{"I think [robot] should introduce itself to my friends because then it's more of a friend to us all"} \\
K & P4 & R $\rightarrow$ O & part of convo & \xmark & \textit{"The robot would be like just a piece of adaptive equipment and so not...participating in the conversation"} \\
L & P5 & R $\rightarrow$ O & part of convo & \xmark & \textit{"We're all having a conversation and I would like the robot to chime in the conversation so it doesn't act peculiar"} \\
M & P2 & R $\rightarrow$ O & signalling staff & \xmark & \textit{"Maybe [Robot]'s light would just turn on and it would signal the waitress to come over"} \\
N & P5 & R $\rightarrow$ O & engage kids & \xmark & \textit{"Robot could ask them...`would you like a coloring page and...some crayons'?"} \\
\hline
\end{tabular}
\end{table*}

\begin{table*}[h!]
\centering
\caption{Participants' perspectives on interaction ecology in robot-assisted social dining (Part 2). ``R'' indicates robot, ``U'' indicates user, and ``O'' indicates others (dining companions, staff, caregiver). ``\cmark'' indicates agreement and ``\xmark'' indicates disagreement.}
\label{tab:rq1-acc-table-b}

\begin{tabular}{l l l l p{1cm} p{9cm}}
\hline
\textbf{Lbl.} & \textbf{P. ID} & \textbf{Sub-theme} & \textbf{Task} & \textbf{Agree?} & \textbf{Quote} \\
\hline
O & P4 & U $\rightarrow$ R & neural interface & \xmark & \textit{"You could have...an AI chip that is interactive, that's more intuitive...with thoughts"} \\
P & P3 & U $\rightarrow$ R & finger pad & \xmark & \textit{"Maybe a finger pad that doesn't take a lot of physical movement"} \\
Q & P4 & U $\rightarrow$ R & wand & \xmark & \textit{"I envision the robot taking my instructions either verbal commands or type commands or perhaps through a wand"} \\
R & P2 & U $\rightarrow$ R & written/typed & \xmark & \textit{"[I would] even [pass] a note if [robot] could read and understand...visualize...something that wouldn't interrupt the meeting"} \\
S & P2 & U $\rightarrow$ R & voice & \xmark & \textit{"I would ask my robot if he would mind cutting my food, my meat for me"} \\
T & P2 & U $\rightarrow$ R & buttons & \xmark & \textit{"I would...push button...for help"} \\
U & P4 & U $\rightarrow$ R & sony controller & \xmark & \textit{"You could have...Sony game controller for a robot"} \\
V & P5 & U $\rightarrow$ R & laptop/tablet & \xmark & \textit{"Maybe I could type it on my iPad and have [the robot] read"} \\
\hline
\end{tabular}
\end{table*}

\begin{table*}[h!]
\caption{Participants' perspectives on context-sensitive robot behavior in social dining}
\label{tab:participants-acc-rq2}
\centering
\begin{tabular}{l l p{5cm} p{8cm}}
\hline
\textbf{Lbl.} & \textbf{P. ID} & \textbf{Sub-theme} & \textbf{Quote} \\
\hline

A & P5 & fade into the background &
\textit{"[Robot looks] as normal as possible"} \\

B & P4 & fade into the background &
\textit{"[Robot] would be...just a piece of adaptive equipment and...not...participating in the conversation or really...a discussion topic"} \\

C & P2 & fade into the background &
\textit{"I don't want [the robot] to draw a lot of attention."} \\

D & P5 & social etiquette &
\textit{(When getting food from shareable dish): "The robot should say to Megan why don't you go ahead and get some, then I'll help and get some food"} \\

E & P3 & support, but not initiate, in shared table practices &
\textit{"[Robot] would...be small and inconspicuous not taking up a lot of space around me or interfering in the space of my dining companions"} \\

F & P5 & personality needs to be versatile and adaptable &
\textit{"Jackie, Megan and I are in the restaurant, the robot just introduced herself as Claire."} \\

G & P3 & personality needs to be versatile and adaptable &
\textit{"I imagine it just being kind of off in the corner and shy and not in the forefront unless it's needed."} \\

H & P6 & personality needs to be versatile and adaptable &
\textit{"I would want [the robot] to understand the quirkiness that it is...it's not human but it's there to act like a human in kind of a way"} \\

I & P1 & pre-programming &
\textit{"I would have to...program that to do all those actions (passing/sharing)...[using] my voice"} \\

J & P4 & pre-programming &
\textit{"The robot should be pre-programmed by the person who is disabled...so that it can be prepared for what the person needs from it."} \\

K & P5 & assess the user + check in &
\textit{"If we start laughing, I want the robot to be able to detect when we're laughing so that she (robot) doesn't make me choke."} \\

L & P2 & assess the user + check in &
\textit{"If I'm having a bad day, [robot] probably already knows that. I've checked in at the beginning telling [the robot] my hands are sore"} \\

M & P5 & assess the user + check in &
\textit{"Maybe have Claire (robot) keep an eye on when my friends are eating so that she (robot) can help me eat when they take bites."} \\

N & P4 & moving across space &
\textit{"[Robot] would move...on wheels...if we are sharing...and the food is in the middle of the table, not everything's going to be...within reach. [Robot] may have to be able to drive itself around other people...to get the curry or rice."} \\

O & P3 & aware of surroundings &
\textit{"[Robot] could...visually scan the area to recognize the different items to...know this is a plate, this is a bowl, this is a cup"} \\

\hline
\end{tabular}

\Description[Participant thoughts on context-sensitive robot behavior in social dining]
{This table captures participant quotes supporting different subthemes that are highlighted in Section~\ref{RQ2-results}. The subthemes are: (1) fade into the background, (2) social etiquette, (3) help, but not initiate, in shared table practices, (4) personality needs to be versatile and adaptable, (5) pre-programming, (6) assess the user + check in, (7) moving across space, and (8) aware of surroundings.}
\end{table*}

\begin{table*}[t]
\caption{Participants' viewpoints on robot's role during and outside of mealtime}
\label{tab:participants-acc-role-rq3}
\centering
\begin{tabular}{l l l p{4cm} p{7cm}}
\hline
\textbf{Lbl.} & \textbf{P. ID} & \textbf{Sub-theme} & \textbf{Robot Role} & \textbf{Quote} \\
\hline

A & P4 & During Mealtime & self-serve &
\textit{"If it is a computer menu like on a tablet and the person is disabled and they don't feel comfortable with technology, maybe the robot could work the tablet for the person [to order]"} \\

B & P2 & During Mealtime & self-serve &
\textit{When ordering, "I might forget if I can't see the menu very well so if the robot has a repeat button maybe where [it] could read back what we finalized that would help me"} \\

C & P5 & During Mealtime & self-serve &
\textit{"I tell the robot: will you please help me look at the menu and help me figure out what I want to eat at the restaurant because it's all vegan here and its gonna be hard for me to decide"} \\

D & P5 & During \& Outside Mealtime & -- &
\textit{"I [ask] the robot, would you please get my medicine out and help me to put it in my mouth so I don't drop it on the floor"} \\

E & P4 & During Mealtime & sharing \& passing &
\textit{"waiter brings a shared dish but I can't reach so I wonder if...your [robot arm] is either an extension or it's folded where you can open it and reach further...I might also need [robot] to pass it around"} \\

F & P1 & During Mealtime & sharing \& passing &
\textit{"I ask one of my family members to pass the appetizers to me and then have my robot hand grasp one of the items"} \\

G & P4 & During Mealtime & sharing \& passing &
\textit{"There's a shared dish in the middle of the table and I would like more of that,...[then] the robot [can] be signalled by myself, then be able to reach into the middle of the table...and put some on my plate which is the correct amount that I want"} \\

H & P4 & Outside Mealtime & support with navigation &
\textit{"[Robot should] be able to grab my walker because I like walking on my own...and helping me with that...I also could see it becoming...a mobility assistance itself"} \\

I & P2 & Outside Mealtime & support with navigation &
\textit{"[Robot could] direct to the bathroom because...even if I can walk there...I still don't want to walk all over the restaurant. I still want to have...a direct path"} \\

J & P3 & Outside Mealtime & support with navigation &
\textit{"[Robot could] maybe pull out a chair"} \\

K & P2 & Outside Mealtime & support with navigation &
\textit{"I would want to use my cane...maybe the robot can carry...the bill and...purse up to the front"} \\

L & P3 & Outside Mealtime & support with navigation &
\textit{"[Robot can] move things around so that I can get to the table"} \\

M & P4 & During \& Outside Mealtime & -- &
\textit{"The robot [should] gather up some things like...put the eating utensils on the plate and prepare to throw away some of the napkins or different things that are trash...just to organize the dining area for the person with disabilities"} \\

N & P4 & Outside Mealtime & payment &
\textit{"Robots could have a payment interface chip where when it's time to pay for the meal, the robot could be activated through...credit card...to pay for the meal...Sometimes it feels like I'm fumbling through my purse..."} \\

O & P2 & Outside Mealtime & payment &
\textit{"[Robot] could use [its] arm to pull up my purse and [it] could even open the zipper to pull out the wallet"} \\

P & P3 & Outside Mealtime & payment &
\textit{"Robot would be able to pick up the receipt so that I could read how much my part of the bill is. If it is in one of those little folders..., it would have the capability to open the folder so it could pull the receipt out"} \\

\hline
\end{tabular}

\Description[Participant thoughts on robot role during and outside of mealtime]
{This table captures participants' thoughts on robot role during and outside of mealtime. There are participant quotes that support different subthemes (during mealtime, outside of mealtime, and during+outside of mealtime) and various robot roles (self-serve, sharing \& passing, support with navigation, payment, and miscellaneous).}
\end{table*}

\end{document}